\documentclass[10pt,twocolumn,letterpaper]{article}

\usepackage[pagenumbers]{cvpr} 

\usepackage{amsmath} 
\usepackage{booktabs} 
\usepackage{xcolor} 
\usepackage{multirow}
\usepackage{makecell}
\usepackage{arydshln}  
\usepackage{balance}

\newcommand{\error}{\raisebox{-0.1em}{\includegraphics[width=3em,height=0.8em]{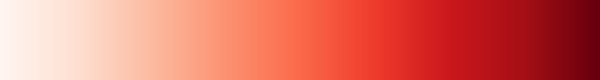}}}
\newcommand{\depth}{\raisebox{-0.1em}{\includegraphics[width=3em,height=0.8em]{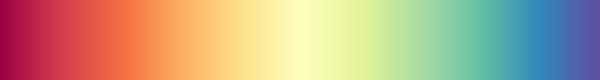}}}

\newcommand{\method}{CAPA}

\newcommand{\Depth}{\mathbf{D}}
\newcommand{\DepthRaw}{\mathbf{d}}
\newcommand{\Image}{\mathbf{I}}
\newcommand{\Mask}{\mathbf{M}}
\newcommand{\Condition}{\mathbf{C}}
\newcommand{\FModel}{\mathbb{F}}
\newcommand{\Param}{\theta}

\newcommand{\Query}{\mathbf{Q}}
\newcommand{\Key}{\mathbf{K}}
\newcommand{\Value}{\mathbf{V}}

\newcommand{\PQ}{\mathbf{W}_q}
\newcommand{\PK}{\mathbf{W}_k}
\newcommand{\PV}{\mathbf{W}_v}
\newcommand{\Seq}{\mathbf{X}}

\definecolor{cvprblue}{rgb}{0.21,0.49,0.74}
\usepackage[pagebackref,breaklinks,colorlinks,allcolors=cvprblue]{hyperref}
\usepackage{enumitem}

\def\paperID{0} 
\def\confName{CVPR}
\def\confYear{2026}

\title{Depth Completion as Parameter-Efficient Test-Time Adaptation}

\newcommand{\nvidia}[0]{\textsuperscript{1}}
\newcommand{\ethz}[0]{\textsuperscript{2}}
\vspace{-0.2em}
\author{
Bingxin Ke\nvidia\textsuperscript{,}\ethz \quad
Qunjie Zhou\nvidia \quad
Jiahui Huang\nvidia \quad
Xuanchi Ren\nvidia \quad
Tianchang Shen\nvidia \quad \\
Konrad Schindler\ethz \quad
Laura Leal-Taixé\nvidia \quad
Shengyu Huang\nvidia \quad \\
\vspace{2mm}
\nvidia NVIDIA \quad  \ethz ETH Zürich
}

\begin{document}

\twocolumn[{%
\renewcommand\twocolumn[1][]{#1}%

\maketitle

\vspace{-1.4em}
\centering
  \includegraphics[width=\textwidth]{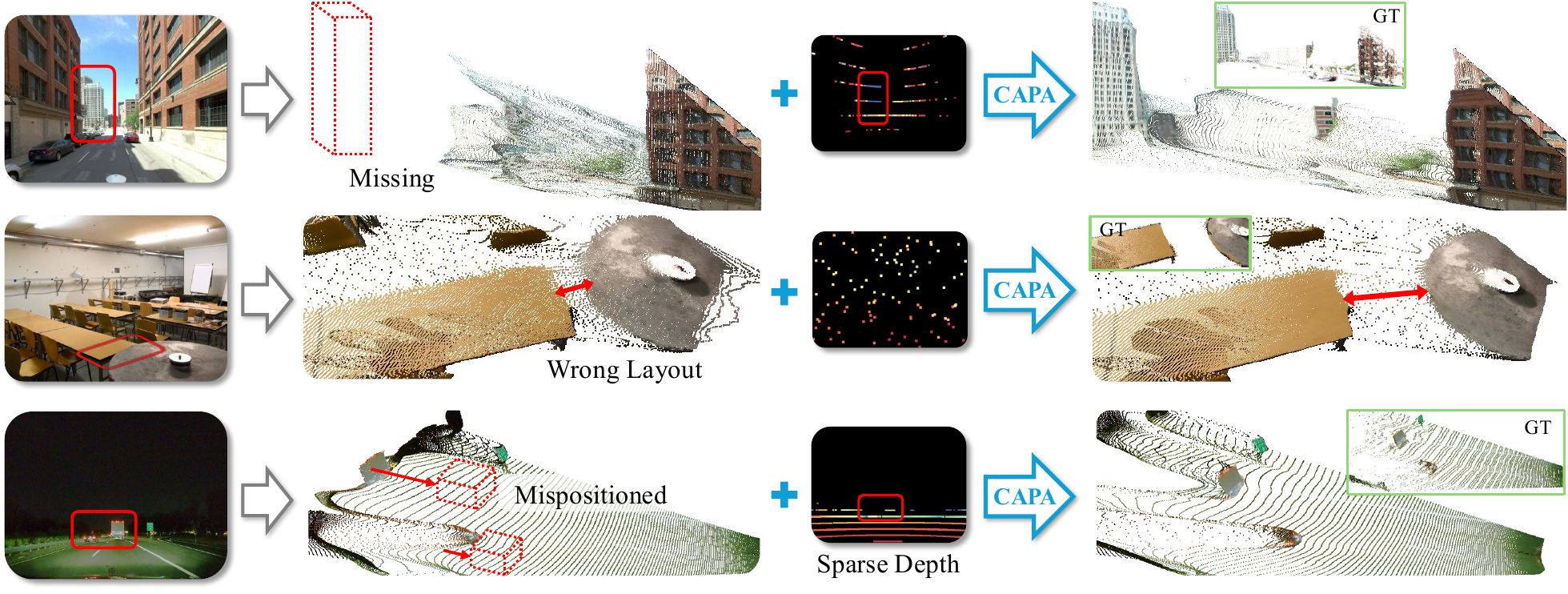}
  \vspace{-1.7em}
  \captionsetup{type=figure}
  \captionof{figure}{
  \textbf{\method{} performs depth completion by adapting geometric foundation models at test-time.} By aligning the strong geometric prior of a base model with the sparse depth information of test samples, one obtains accurate reconstructions of scene layout and fine details, overcoming limitations of the base model such as distorted surfaces and misplaced objects, even under challenging conditions.
  }
  \label{fig:teaser}
 \vspace{1.8em}
 }
]

\begin{abstract}
\vspace{-0.2em}
We introduce \method{}, a parameter-efficient test-time optimization framework that adapts pre-trained 3D foundation models (FMs) for depth completion, using sparse geometric cues. Unlike prior methods that train task-specific encoders for auxiliary inputs, which often overfit and generalize poorly, \method{} freezes the FM backbone. Instead, it updates only a minimal set of parameters using Parameter-Efficient Fine-Tuning (\eg LoRA or VPT), guided by gradients calculated directly from the sparse observations available at inference time. This approach effectively grounds the foundation model's geometric prior in the scene-specific measurements, correcting distortions and misplaced structures. For videos, \method{} introduces sequence-level parameter sharing, jointly adapting all frames to exploit temporal correlations, improve robustness, and enforce multi-frame consistency. \method{} is model-agnostic, compatible with any ViT-based FM, and achieves state-of-the-art results across diverse condition patterns on both indoor and outdoor datasets. 
Project page: \href{https://research.nvidia.com/labs/dvl/projects/capa}{research.nvidia.com/labs/dvl/projects/capa}.

\end{abstract}


\section{Introduction}
Monocular depth estimation is an elementary, yet fundamentally ill-posed capability of a vision system. The advent of large, pre-trained models for spatial perception, often termed \emph{3D foundation models (FMs)}~\cite{wang2025vggt,yang2024depthv2}, has redefined this task. These models, trained on vast amounts of data, capture rich geometric and semantic priors and achieve impressive zero-shot generalization across diverse scenes. But even the most capable FMs are not omnipotent: their reconstructions may exhibit defects such as incorrect scaling and spatially varying affine distortions, particularly when deployed in previously unseen environments or imaging conditions. 
For reliable and metrically accurate results, the visual systems must be grounded using external \textit{geometric cues}, typically accurate yet sparse observations, \eg, from LiDAR or structure-from-motion (SfM). 
Therefore, the general problem of \textit{depth completion}, \ie, recovering dense metric geometry conditioned on sparse constraints, remains a fundamental requirement.

Traditional depth completion methods~\cite{wang2025depthprior, zuo2025omni} address the problem by designing and training task-specific networks that explicitly use the sparse depth map as an additional input channel. This approach suffers from two key limitations. First, it is practically impossible to curate datasets that capture all possible geometric configurations and sparsity patterns. Consequently, the encoder in such models tends to overfit to the training configurations and may fail to generalize to unfamiliar input patterns---sometimes performing worse than unconstrained monocular estimators (\cf \cref{tab:main_result}).
Second, retraining (respectively, fine-tuning) a foundation model for each new setup is computationally expensive and risks degrading the pre-trained, generic representation that underpins its broad applicability.

We approach the problem from a different perspective. Rather than (re-)training a model offline on a curated dataset that incorporates sparse 3D observations as additional inputs, we perform test-time adaptation for each sample individually. This allows us to retain the broad geometric prior already encoded in the base model while guiding it towards the sparse evidence available at inference. To this end, we formulate depth \textbf{c}ompletion \textbf{a}s \textbf{p}arameter-efficient test-time \textbf{a}daptation (\method{}) of a frozen 3D foundation model.

To practically realize per-sample adaptation at test-time, we leverage Parameter-Efficient Fine-Tuning (PEFT) \cite{hu2022lora, jia2022visual}, rather than naively fine-tuning the entire model. PEFT methods update only a compact set of parameters while keeping the main backbone frozen, significantly enhancing computational and memory efficiency while maintaining high accuracy. In our formulation, the sparse depth map provides the necessary gradient guidance to update these lightweight PEFT components, such as low-rank adapters (LoRA \cite{hu2022lora}) or learnable prompt tokens (VPT \cite{jia2022visual}). This strategic approach enables efficient, scene-specific adaptation at inference while preserving the global geometric understanding encoded within the original foundation model.

We further extend \method{} to the video domain. The PEFT formulation naturally supports sharing a common set of learnable parameters across frames, enabling sequence-level adaptation. Instead of tuning each frame independently, we jointly optimize a small shared parameter set over the entire sequence, leveraging the strong geometric and appearance correlations between neighboring frames. This aggregation of depth cues improves robustness to sparse or noisy observations and enforces temporal consistency (\cf~\cref{tab:token_share}). In practice, adaptation is performed efficiently with mini-batch updates, yielding a compact, scene-consistent calibration for all frames.

\method{} is broadly compatible and model-agnostic, applicable to any 3D foundation model with a Vision Transformer (ViT) backbone. By aligning the base model’s geometric prior with the sparse depth of each test sample, \method{}  produces accurate, dense reconstructions that adapt to scene-specific geometry and correct inherent distortions (\cf~\cref{fig:teaser}). Through extensive experiments on diverse datasets, \method{} consistently achieves state-of-the-art performance with significantly lower errors than prior methods. Qualitatively (\cf~\cref{fig:qualitative_main}), this yields cleaner, more coherent depth maps and finer structural details, underscoring \method{}'s strength as a practical and effective way to specialize foundation models to new scenes directly at inference time.
This capability effectively targets high-fidelity offline applications, such as 3D mapping and pseudo-ground-truth generation.
In summary, our contributions are:
\begin{itemize}
\item reframing depth completion as \emph{parameter-efficient adaptation} of foundational 3D vision models, guided by sparse depth available at test time;
\item introducing \emph{sequence-level parameter sharing} for video depth completion, leveraging strong inter-frame correlations for improved temporal consistency and robustness;
\item conducting comprehensive evaluation and ablation studies with two PEFT strategies (LoRA and VPT) and three distinct base models (VGGT, MoGe-2, UniDepthV2);
\item achieving the best performance on four datasets captured in diverse environments, with varying condition patterns.
\end{itemize}

\section{Related Work}
\label{sec:realated}

\subsection{Depth estimation}
Monocular depth estimation is a dense regression problem that has advanced rapidly over the past decade. Early work by Eigen et al.~\cite{eigen_depth_2014} laid the foundation, followed by methods exploring ordinal regression~\cite{fu2018dorn}, depth binning~\cite{Farooq_Bhat_2021, bhat2023zoedepth}, canonical camera-space representations~\cite{yin2023metric3d}, \etc.
A major step forward was improving generalization through training on large-scale datasets~\cite{Ranftl2020_midas, ranftl2021_dpt}. More recently, foundation models trained on massive and diverse data~\cite{yang2024depthv1, yang2024depthv2, wang2025moge, wang2025moge2} or repurposed from diffusion-based image generators~\cite{ke2023repurposing, ke2025marigold, fu2024geowizard} have achieved state-of-the-art zero-shot performance across a wide range of scenes.

The foundation model paradigm has also extended to video depth estimation. Earlier approaches primarily relied on optical flow and pose optimization to enforce temporal consistency~\cite{Luo-VideoDepth-2020, zhang2021consistent}. More recent methods, built upon strong pre-trained priors~\cite{rombach2022high,oquab2023dinov2}, such as RollingDepth~\cite{ke2025video} and VideoDA~\cite{chen2025video}, achieve robust generalization and temporal coherence without the need for explicit post-processing. 

In parallel, feed-forward 3D reconstruction models have emerged, jointly predicting multiple geometric outputs. This line of work was pioneered by DUSt3R~\cite{wang2024dust3r}, which predicts point maps from unposed image pairs, requiring post-processing to extract camera parameters. Methods like VGGT~\cite{wang2025vggt}, Fast3R~\cite{Yang_2025_Fast3R}, and $\pi^3$~\cite{wang2025pi} extended this to the multi-view setting, leveraging large-scale multi-task training to directly predict depth, camera poses, and intrinsics. Subsequent work, such as MapAnything~\cite{keetha2025mapanything}, further enhanced this direction by conditioning on available geometric cues. Our proposed \method{} builds upon these strong geometric foundation models, enhancing them through our test-time adaptation framework.


\subsection{Parameter-Efficient Fine-Tuning}
The vast scale of Foundation Models has driven the development of Parameter-Efficient Fine-Tuning (PEFT) methods, which adapt large models by freezing the backbone and updating only a small, additive or reparameterized set of parameters~\cite{han2024parameterefficient}. 
We focus on two primary types when adapting attention~\cite{vaswani2017attention} layers: Low-Rank Adaptation (LoRA)~\cite{hu2022lora}, which updates projection matrices via low-rank decomposition, and Visual Prompt Tuning (VPT)~\cite{jia2022visual}, which injects learnable tokens into the input token sequence. The technical details are provided in \cref{sec:peft}.

In addition to LLM tasks, PEFT has recently been explored in the 3D domain. Prior works such as LoRA3D~\cite{lu2024lora3d} and Test3R~\cite{yuan2025test3r} apply PEFT to 3D reconstruction, focusing on scene-specific calibration using consistency loss across frame triplets. In this work, we investigate PEFT for depth completion, assuming the availability of sparse geometric cues, and frame the task as test-time adaptation of pre-trained depth models.

\subsection{Depth Completion}
Early methods propagated sparse measurements via spatial propagation networks (SPNs)~\cite{cheng2019learning, cheng2020cspn++, park2020non}, affinity kernels~\cite{imran2019depth}, or bilateral propagation~\cite{tang2024bilateral}. Distinct from pure learning, other works explicitly grounded completion in geometry by modeling topology via scaffolding~\cite{wong2020unsupervised} or enforcing intrinsic constraints through calibrated backprojection layers~\cite{wong2021unsupervised}. Later approaches captured non-local structure using graph neural networks~\cite{xiong2020sparse}, hybrid CNN-Transformer~\cite{zhang2023completionformer}, or recurrent frameworks OGNI-DC~\cite{zuo2024ogni} and its multi-scale variant OMNI-DC~\cite{zuo2025omni}.

Recent training-based strategies leverage pre-trained priors by fusing features with sparse depth (\eg, PriorDA~\cite{wang2025depthprior} and PromptDA~\cite{lin2024prompting}) or adapting diffusion models~\cite{gui2025depthfm}. Attempts to achieve sensor-agnostic capabilities include MapAnything~\cite{keetha2025mapanything}, which trains a unified multi-task model, and UniDC~\cite{park2024simple}, which explores hyperbolic embeddings.

Finally, test-time adaptation (TTA) methods refine predictions directly at inference. Diffusion-based approaches such as Marigold-DC~\cite{viola2024marigold} and related works~\cite{hyoseok2025zero,gregorek2025steeredmarigold} update latent features using gradients from the sparse supervision. 
Other recent approaches adapt pre-trained depth completion models using sparse depth proxies~\cite{park2024test}, learned energy functions~\cite{chung2025eta}, or domain-specific embeddings for unsupervised continual adaptation~\cite{Rim_2025_CVPR}.
A concurrent effort, TestPromptDC~\cite{jeong2025test}, pursues a similar adaptation goal but operates in image space with pixel-level prompts. This design is highly sensitive to noise and not well-suited for video, as it lacks an effective mechanism for parameter sharing. In contrast, our framework enables robust scene-specific adaptation precisely by sharing visual prompts or low-rank adapters across multi-frame sequences, a key capability for video consistency missing in prior TTA works.

\begin{figure}[t]
    \centering
    \includegraphics[width=\linewidth]{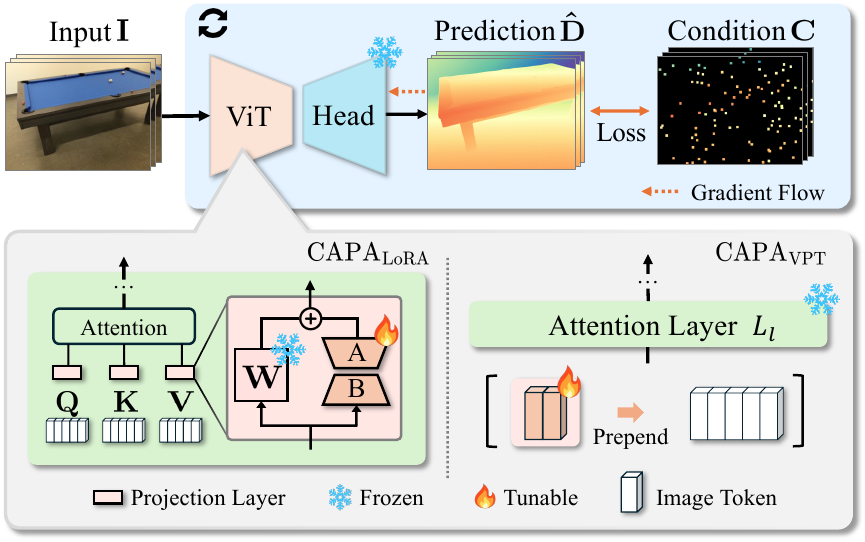}
    \caption{\textbf{Method overview of \method{}.} 
    \method{} adapts 3D foundation models given sparse conditional depth ($\Condition$) by efficiently tuning its image encoder while keeping all pre-trained weights frozen. This is achieved by manipulating the attention layers via two methods: 1) \method{}$_\text{LoRA}$, which adds low-rank adapters to the projection weights, or 2) \method{}$_\text{VPT}$, which prepends tunable prompt tokens to the image token sequence before each attention layer.
    }
    \label{fig:method}
    \vspace{-0.8em}
\end{figure}

\section{Method} \label{sec:method}
\paragraph{Problem statement.}
Given an input RGB image $\Image \in \mathbb{R}^{H \times W \times 3}$ and a corresponding sparse depth map $\Condition \in \mathbb{R}^{H \times W}$ (where valid measurements are indicated by a mask $\Mask \in \mathbb{R}^{H \times W}$), \textit{depth completion} is to predict the complete, dense depth map $\hat{\Depth} \in \mathbb{R}^{H \times W}$ that is consistent with the sparse input cues.
We further extend this to the multi-view or video setting, where $N$ frames and depth maps are jointly used to produce geometrically and temporally consistent predictions $(\hat{\Depth}_1,\dots,\hat{\Depth}_N)$.

\subsection{VGGT Recap}
VGGT~\cite{wang2025vggt} is a 3D foundation model that jointly infers camera intrinsics, poses, and depth from unposed image sets. It employs a Vision Transformer (ViT) encoder~\cite{oquab2023dinov2} composed of multiple attention blocks,
which extract patch tokens representing local image features. 
These tokens, together with register and camera tokens, are processed by a multi-view aggregator 
that propagate geometric information by alternating between self-attention within each frame and global attention across all frames.
The aggregated features are then decoded into depth maps and point maps by DPT~\cite{ranftl2021_dpt} heads, and camera parameters by a camera head. In this paper, we only utilize the ViT encoder, aggregator, and the depth head.
While we adopt VGGT for its strong geometric priors, our \method{} framework is broadly applicable to other ViT-based 3D foundation models (\cf~\cref{fig:other_model}).

\subsection{Parameter-Efficient Fine-Tuning}
\label{sec:peft}
In this section, we introduce two complementary PEFT techniques designed to change the final attention outputs with a minimal number of learnable parameters. We first recap the core attention mechanism~\cite{vaswani2017attention}, focusing on the scaled dot-product attention formulation for clarity.

Given an input token sequence $\Seq \in \mathbb{R}^{L \times d_c}$, where $L$ is the token length and $d_c$ the channel dimension, the attention update is:
\vspace{-0.5em}
\begin{equation}\label{eq:attention}
\text{Attention}(\Seq) = \text{Softmax}\left( \frac{\Query' \Key'^T}{\sqrt{d_k}} \right)\Value',
\end{equation}
where queries, keys, and values are obtained via learned projections $\PQ, \PK, \PV \in \mathbb{R}^{d_c \times d_k}$ ($d_k$ being the projected channel dimension):
\[
\Query' = \Seq \PQ, \quad \Key' = \Seq \PK, \quad \Value' = \Seq \PV.
\]

\vspace{-1.5em}
\paragraph{Low-rank adaptation.}
LoRA~\cite{hu2022lora} adapts the model by augmenting the projection matrices $\mathbf{W}_m$ ($m \in \{q, k, v\}$) with low-rank updates. The key assumption is that the task-specific weight update $\Delta \mathbf{W}_m$ lies in a low-dimensional subspace:
\begin{equation}\label{eq:lora}
\mathbf{W}'_m = \mathbf{W}_m + \Delta \mathbf{W}_m, \quad 
\Delta \mathbf{W}_m = \mathbf{B}_{m}\mathbf{A}_{m},
\end{equation}
where $\mathbf{B}_{m}\!\in\!\mathbb{R}^{d_c \times r}$ and $\mathbf{A}_{m}\!\in\!\mathbb{R}^{r \times d_k}$ and $r\!\ll\!\min(d_c, d_k)$. 
During adaptation, only $\mathbf{A}_{m}$ and $\mathbf{B}_{m,}$ are trained, while $\mathbf{W}_m$ remain frozen. This enables LoRA to alter attention projections through a compact low-rank update, efficiently adjusting the attention response.

\vspace{-0.5em}
\paragraph{Visual prompt tuning.}
VPT~\cite{jia2022visual} instead modulates the attention computation by expanding the token sequence. At each transformer layer, a small set of trainable \emph{prompt tokens} $\mathbf{P} \in \mathbb{R}^{t \times d_c}$ is prepended to the input sequence:
\begin{equation}\label{eq:vpt_concat}
\Seq_{\text{new}} = [\mathbf{P}; \Seq].
\end{equation}
The attention is then computed over the extended token sequence, where the trainable prompt tokens modulate the attention map and thus influence how image tokens attend to one another. After the attention layer, only the updated image tokens are retained and passed forward, preserving the original token lengths.

\vspace{0.5em}
\textbf{In summary}, both LoRA and VPT achieve parameter-efficient model adaptation by manipulating the same component---attention. LoRA modifies the \emph{projection layers} to alter the projected token values directly, whereas VPT injects \emph{learnable tokens} to reshape the attention distribution implicitly. While they offer two complementary mechanisms, their effect is similar: modulate the attention to customize a pre-trained model with task-specific cues.

\begin{table*}[t]
    \centering
    \caption{\textbf{Quantitative comparison of \method{} with baseline methods}
    [AbsRel (\%)$\downarrow$].
    }
    \label{tab:main_result}
    \resizebox{\linewidth}{!}{
        \setlength{\tabcolsep}{6pt} 

\begin{tabular}{l *{12}{r} r}
\toprule
 & \multicolumn{3}{c}{ScanNet} & \multicolumn{3}{c}{7-Scenes} & \multicolumn{3}{c}{iBims} & \multicolumn{3}{c}{Metropolis} & \multirow{2}{*}{\makecell{Avg.\\Rank}} \\
 \cmidrule(lr){2-4} \cmidrule(lr){5-7} \cmidrule(lr){8-10} \cmidrule(lr){11-13}
 & SIFT & 100 & $<3m$ & SfM & 100 & $<3m$ & SIFT & 100 & $<5m$ & 8-line & 16-line & 32-line & \\

\midrule

DepthAnythingV2~\cite{yang2024depthv2} & 4.8 & 3.8 & 4.7 & 5.4 & 4.6 & 4.6 & 4.0 & 3.3 & 9.2 & 54.0 & 57.0 & 56.2 & 9.8 \\
UniDepthV2~\cite{piccinelli2025unidepthv2} & 3.2 & 2.7 & 2.8 & 4.2 & 3.7 & 3.7 & 3.4 & 2.7 & 3.2 & 29.4 & 28.3 & 27.7  & 6.6\\
MoGe-2~\cite{wang2025moge2} & 4.1 & 3.3 & 3.4 & 3.9 & 3.3 & 3.3 & 3.4 & 2.7 & \underline{3.1} & 22.1 & 20.7 & 20.2 & 6.1 \\
VGGT~\cite{wang2025vggt} & 2.2 & 2.0 & 2.0 & 4.7 & 4.2 & 4.2 & 4.0 & 3.3 & 3.9 & 10.2 & 10.2 & 9.9 & 5.5 \\
VideoDA~\cite{chen2025video} & 5.0 & 4.1 & 4.2 & 6.1 & 5.1 & 5.1 & 5.2 & 4.0 & 4.6 & 50.8 & 47.3 & 46.5 & 10.0 \\
\noalign{\vskip 0.6ex}\cdashline{1-14}\noalign{\vskip 0.6ex}
PromptDA~\cite{lin2024prompting} & 16.6 & 13.2 & 12.2 & 8.2 & 6.9 & 4.0 & 11.6 & 7.8 & 8.0 & 27.9 & 19.6 & 16.5 & 10.1 \\
PriorDA-v1.1~\cite{wang2025depthprior} & 3.4 & 2.5 & 10.3 & 3.7 & 3.1 & 9.1 & 2.9 & 2.6 & 14.9 & 26.3 & 19.5 & 15.6 & 6.8 \\
OMNI-DC-v1.1~\cite{zuo2025omni} & 2.0 & 1.5 & 5.4 & 3.3 & 2.2 & 4.6 & \underline{1.7} & \textbf{1.5} & 10.8 & 14.5 & 11.4 & 9.1 & 4.8 \\
Marigold-DC~\cite{viola2024marigold} & 4.3 & 3.7 & 2.2 & 4.1 & 4.0 & 1.6 & 7.6 & 6.1 & 5.2 & 34.0 & 25.4 & 20.0 & 7.8 \\
TestPromptDC~\cite{jeong2025test} & 4.4 & 3.8 & 2.8 & 3.1 & 4.0 & 2.6 & 7.0 & 5.1 & 5.1 & 21.8 & 18.9 & 16.9 & 6.8 \\
\midrule
\method{}$_\text{LoRA}$ & \textbf{1.0} & \textbf{0.9} & \underline{1.1} & \underline{2.8} & \textbf{1.0} & \textbf{0.9} & \underline{1.7} & 2.1 & \textbf{2.0} & \textbf{7.8} & \textbf{7.1} & \textbf{6.7} & \textbf{1.4} \\
\method{}$_\text{VPT}$ & \underline{1.1} & \underline{1.0} & \textbf{1.0} & \textbf{2.6} & \underline{1.1} & \underline{1.1} & \textbf{1.4} & \underline{1.7} & \textbf{2.0} & \underline{8.2} & \underline{7.6} & \underline{7.4} & \underline{1.7} \\
\bottomrule


\end{tabular}

    }
\end{table*}

\subsection{Depth Completion as Model Adaptation}
\paragraph{Model adaptation.} 
Given an input pair $(\Image, \Condition)$, we adapt the pre-trained model $\FModel$ to sparse depth measurements by iteratively optimizing a minimal set of parameters $\Param$, corresponding to either LoRA's low-rank projection matrices $(\mathbf{A}_{m}, \mathbf{B}_{m})$ or VPT's prompt tokens $\mathbf{P}$.

At each iteration, the adapted model $\FModel_\Param$ predicts a dense depth map $\hat{\DepthRaw}$. To resolve the scale ambiguity inherent in foundation models, we first compute an affine transformation (scale $s$ and shift $t$) that aligns the raw prediction $\hat{\DepthRaw}$ with the sparse measurements $\Condition$ by solving the robust L1 minimization: 
\begin{equation}
    \min_{s, t} \left| \Mask \odot \big(s \cdot \hat{\DepthRaw} + t - \Condition \big) \right|_1 
\end{equation} 
The resulting aligned depth, $\hat{\Depth} = s \cdot \hat{\DepthRaw} + t$, is then used to compute the L1 loss against $\Condition$ at valid pixels, which is subsequently backpropagated to update the parameters $\Param$.

\vspace{-1.0em}
\paragraph{Sequence-level adaptation.} 
For video or multi-frame inputs, we propose a sequence-level adaptation strategy by sharing the same set of trainable parameters $\Param$ across all frames in a sequence. This design is founded on the observation that frames within the same scene exhibit consistent global geometric and texture characteristics. Consequently, the pre-trained foundation model exhibits highly correlated behavior when processing these frames, allowing a single, shared set of $\Param$ to effectively capture the necessary scene-specific adjustments for the entire sequence. This approach aggregates sparse geometric cues from multiple views, leading to two main benefits: it significantly improves robustness against imperfect or noisy condition points, and enhances geometric consistency across multi-view or temporal frames. 
Compared to per-frame optimization, optimizing over diverse viewpoints within the scene stabilizes the optimization process by averaging out noise and local errors present in individual frame gradients, ensuring that the learned $\Param$ represents a consistent scene-specific calibration.
To efficiently handle long sequences, we employ a mini-batch strategy: at each optimization step, a random subset of frames is sampled to compute the loss and update $\Param$. This strategy substantially reduces the computational overhead compared to processing the entire sequence simultaneously at each step.

\section{Experiments}\label{sec:experiment}

\subsection{Experimental Settings}
\paragraph{Evaluation datasets.}
We evaluate our method on four datasets, covering both indoor and outdoor settings. ScanNet~\cite{dai2017scannet} and 7-Scenes~\cite{glocker2013real} are indoor RGB-D video datasets. For ScanNet, we use 100 test sequences, for 7-Scenes all 18 available sequences. From each sequence, we uniformly sample 100 frames within the first 300 ones of the video. iBims~\cite{koch2018evaluation} is a single-image indoor dataset containing 100 challenging RGB-D samples. Mapillary Metropolis~\cite{antequera2020mapillary} is a city-scale outdoor dataset with survey-grade annotations. We use its validation split, which includes 36 sequences with 68 to 96 frames (after discarding one sequence for which the ground-truth depth is partially missing).

\vspace{-0.5em}
\paragraph{Sparse depth patterns.}
We experiment with different point patterns for 3D conditioning, to reflect diverse possible application scenarios. For the \textit{SfM} setting, we sample SIFT~\cite{Lowe2004SIFT} keypoints on ScanNet and iBims, and use SfM points reconstructed by COLMAP~\cite{schonberger2016structure, Brachmann2021ICCV} for 7-Scenes. For the \textit{random} setting, we uniformly sample 100 points to simulate extremely sparse, but evenly distributed guidance. To mimic depth sensors with \emph{limited range}, we follow PriorDA~\cite{wang2025depthprior} and restrict condition points to depths within 3\,m or 5\,m of the camera. For outdoor scenes in Mapillary Metropolis, we simulate automotive LiDAR scanning patterns with 8, 16, and 32 scan lines. In addition, we follow OMNI-DC~\cite{zuo2025omni} and corrupt 10\% of the condition points with noise. For 7-Scenes (SfM), no additional noise is added since the SfM points are already naturally noisy. Further implementation details, as well as results without noise, are given in the appendix.

\vspace{-0.5em}
\paragraph{Evaluation metrics.}
Following~\cite{wang2025depthprior}, we report AbsRel as the main depth metric (MAE and RMSE are in the appendix). 
We measure temporal consistency using the optical flow–based warping loss (OPW)~\cite{wang2022less, ke2025video}.

\vspace{-0.5em}
\paragraph{Baselines.}
\method{}$_\text{LoRA}$ and \method{}$_\text{VPT}$ denote the two variants of our proposed method on top of VGGT~\cite{wang2025vggt} base model. We compare \method{} with a variety of baselines, representing four different types of dense depth estimators: DepthAnythingV2~\cite{yang2024depthv2}, UniDepthV2~\cite{piccinelli2025unidepthv2}, and MoGe-2~\cite{wang2025moge2} for \textit{single-image} depth estimation; VGGT~\cite{wang2025vggt} for \textit{multi-view} reconstruction; VideoDA~\cite{chen2025video} for \textit{video-based} depth estimation; PromptDA~\cite{lin2024prompting}, OMNI-DC~\cite{zuo2025omni}, PriorDA~\cite{wang2025depthprior}, Marigold-DC~\cite{viola2024marigold}, and TestPromptDC~\cite{jeong2025test} for dedicated \textit{depth completion} schemes.
For the first three groups, which do not utilize depth guidance, we align the predictions to the sparse depth observations in postprocessing, on a frame-by-frame basis.

\vspace{-0.5em}
\paragraph{Implementation details.}
We inject learnable parameters into all 24 attention layers of VGGT's ViT encoder. We set the LoRA~\cite{hu2022lora} rank to $r=4$ and the VPT~\cite{jia2022visual} prompt length to $t=16$. Both variants yield a highly compact set of 0.39M trainable parameters. We tune each sample for 100 steps using the AdamW optimizer with a learning rate of $10^{-3}$ for LoRA and $2\times 10^{-4}$ for VPT. We provide more details in the 
appendix.

\begin{figure*}[t]
    \vspace{-1em}
    \centering
    
    \includegraphics[width=0.99\linewidth]{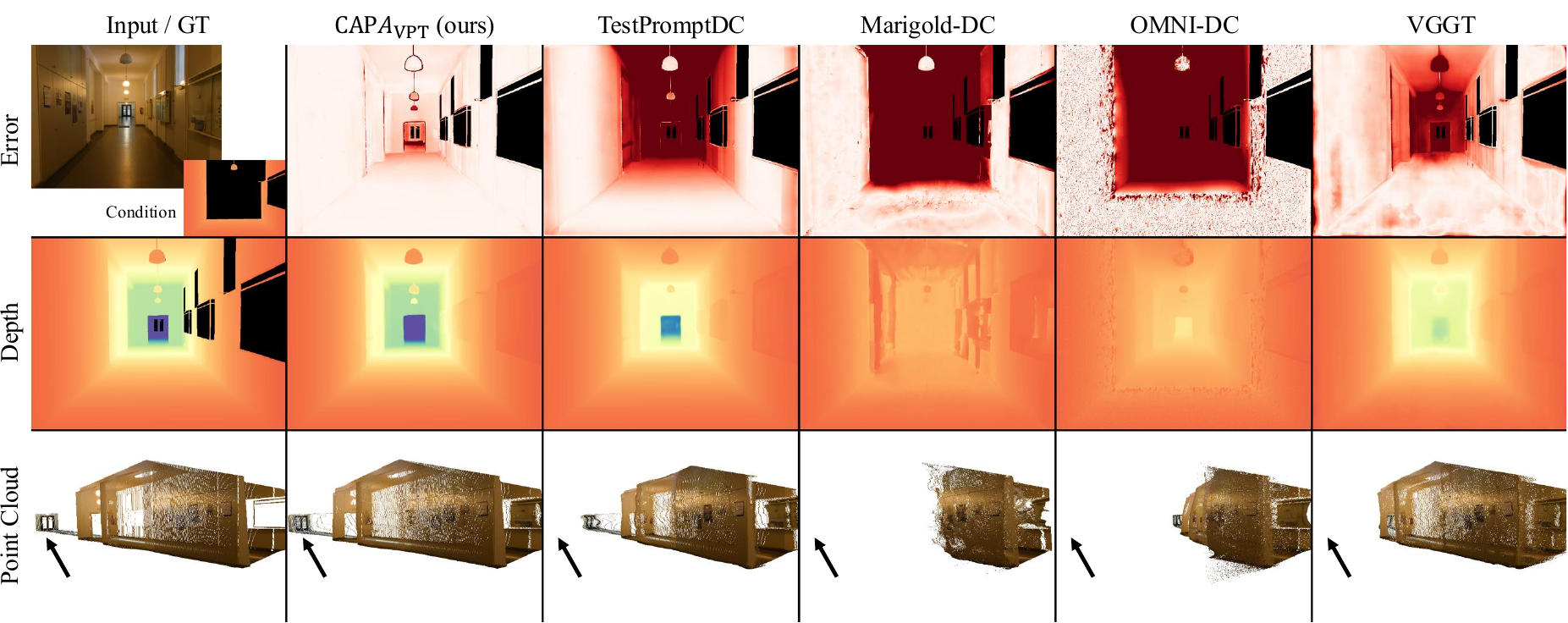}
    \vspace{0.1em} \hrule \vspace{0.1em}
    \includegraphics[width=0.99\linewidth]{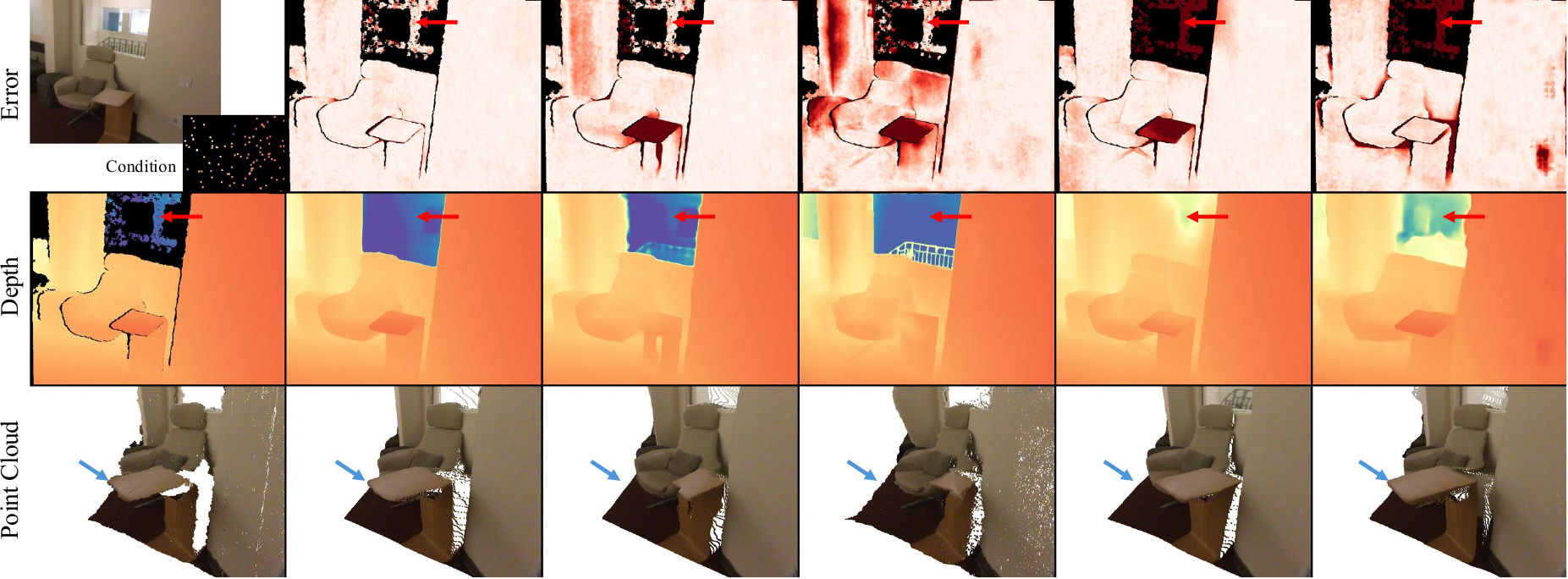}
    \vspace{0.1em} \hrule \vspace{0.1em}
    \includegraphics[width=0.99\linewidth]{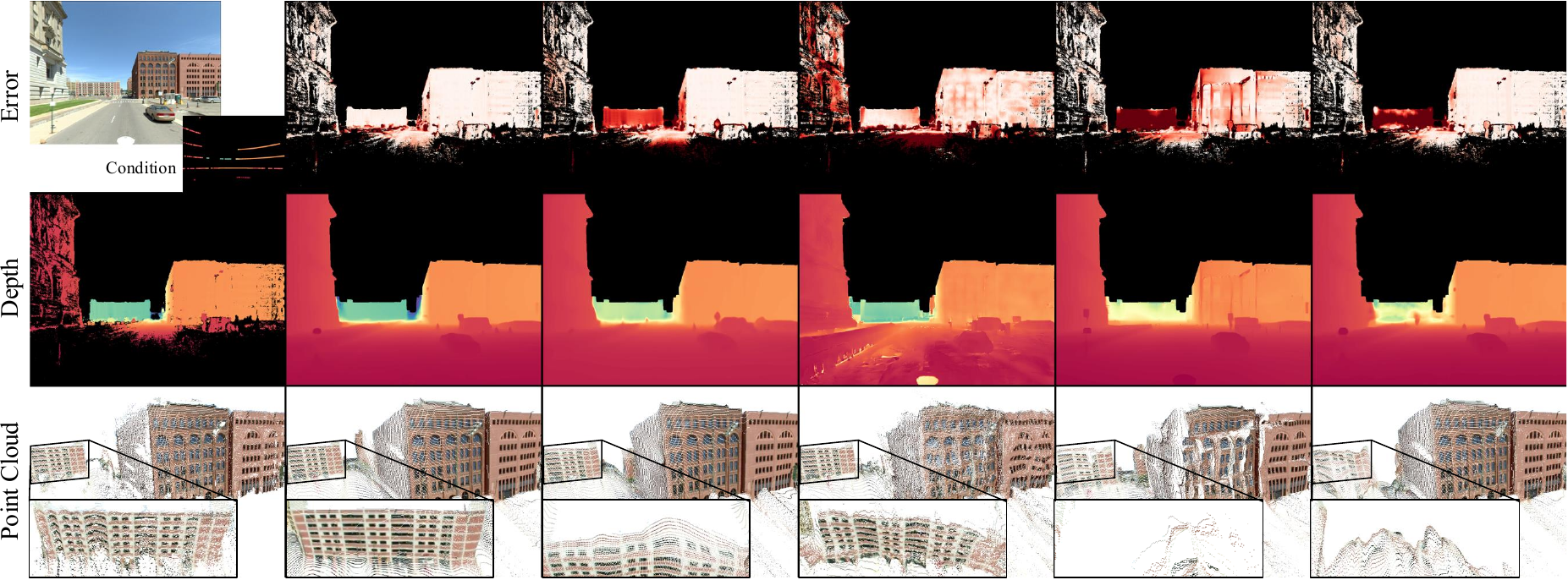}
  
\caption{\textbf{Qualitative comparison on iBims, ScanNet, and Metropolis datasets}. \method{} reliably recovers the full scene: in the first sample, despite 3D points being available only within $<5$\,m, the geometry prior is calibrated sufficiently to reconstruct the full depth range, whereas most baselines focus on the well-constrained near-field; in the second sample, with only two observed points in the far-field, \method{} corrects the global geometry while preserving local structures; in the third sample, geometric structure is correctly recovered both near and far from the camera. Depth is color-coded near~\depth~far, errors low~\error~high. Colored arrows mark corresponding locations across images.}
  \label{fig:qualitative_main}
\end{figure*}

\subsection{Main Results}\label{sec:main_results}

\paragraph{Comparison to state of the art.}
Quantitative results are summarized in~\cref{tab:main_result}.
The error metrics clearly highlight the effectiveness of our parameter-efficient adaptation strategy. The two variants, $\text{\method{}}_\text{LoRA}$ and $\text{\method{}}_\text{VPT}$, achieve comparable performance and consistently outperform all baselines, securing a rank of first or second in every setting across four datasets. Notably, the prediction errors of $\text{\method{}}$ are less than half as large as those of competing depth completion schemes in most cases. Furthermore, adaptation significantly boosts the base model: the VGGT error reduces by 2--3$\times$ when enhanced with \method{}. 

Among depth completion methods, OMNI-DC~\cite{zuo2025omni} comes closest to \method{}, but suffers from a substantial accuracy drop in the limited-range setting, likely because it is not trained for that scenario.  A similar degradation is observed for other methods such as PromptDA~\cite{lin2024prompting} and PriorDA~\cite{wang2025depthprior}. In contrast, test-time adaptation approaches like Marigold-DC~\cite{viola2024marigold} and TestPromptDC~\cite{jeong2025test}, which also adapt pre-trained models at test time, exhibit greater robustness, supporting the effectiveness of foundation models as priors. Qualitative results are displayed in~\cref{fig:qualitative_main}.

\begin{figure}[t]
  \centering
  \includegraphics[width=0.95\linewidth]{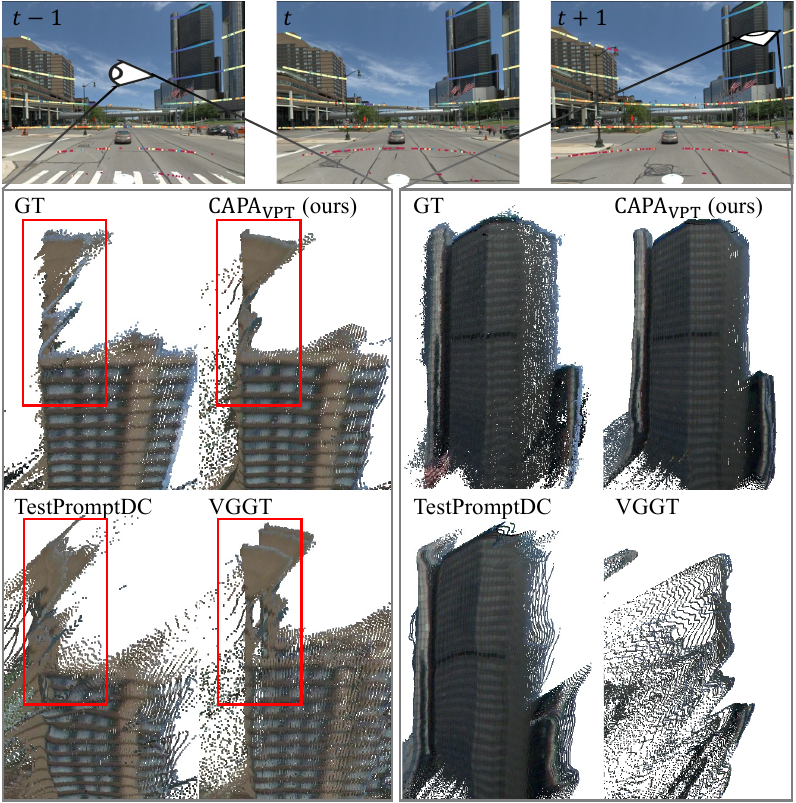}
  \vspace{-0.5em}
    \caption{\textbf{Qualitative comparison of temporal consistency.} Three consecutive frames are \emph{overlaid} using ground truth poses. On the left, \method{} reconstructs coherent building shapes, while the baselines are visibly distorted.
    }
  \label{fig:qualitative_consistency}
\end{figure}
\vspace{-0.6em}

\begin{table}[t]
    \centering
    \caption{\textbf{Temporal consistency evaluation} [OPW(\%)~$\downarrow$].
    }
    \vspace{-0.4em}
    \label{tab:main_opw}
    \resizebox{\linewidth}{!}{
        \setlength{\tabcolsep}{2pt} 

\begin{tabular}{l *{9}{c}}
\toprule
 & \multicolumn{3}{c}{ScanNet} & \multicolumn{3}{c}{7-Scenes} & \multicolumn{3}{c}{Metropolis}  \\
 \cmidrule(lr){2-4} \cmidrule(lr){5-7} \cmidrule(lr){8-10} 
 & SIFT & 100 & $<3m$ & SfM & 100 & $<3m$ & 8-line & 16-line & 32-line \\

\midrule



DepthAnythingV2 & 4.3 & 3.6 & 7.4 & 3.5 & 3.5 & 3.0 & 1296.6 & 1419.9 & 1413.8 \\
UniDepthV2 & 2.7 & 2.5 & 2.5 & \underline{2.9} & \underline{2.9} & 2.6 & 165.2 & \underline{152.2} & 141.1 \\
MoGe-2 & 4.1 & 3.7 & 3.6 & 3.1 & \underline{2.9} & 2.7 & 219.3 & 205.3 & 199.1 \\
VGGT & \underline{2.3} & \underline{2.2} & \underline{2.0} & 3.1 & 3.2 & 2.9 & 149.4 & 152.5 & 149.0 \\
VideoDA & 3.3 & 3.2 & 2.8 & 3.3 & 3.6 & 2.9 & 179.3 & 160.4 & 141.5 \\
\noalign{\vskip 0.6ex}\cdashline{1-10}\noalign{\vskip 0.6ex}
PromptDA & 11.2 & 12.1 & 3.8 & 6.3 & 10.3 & 3.5 & 206.4 & 199.2 & 179.8 \\
PriorDA-v1.1 & 5.0 & 4.7 & 3.7 & 3.5 & 4.9 & 4.1 & 268.9 & 233.2 & 195.0 \\
OMNI-DC-v1.1 & 3.2 & 3.0 & 9.2 & 3.3 & 3.7 & 10.0 & 171.2 & 174.5 & 171.3 \\
Marigold-DC & 8.1 & 9.0 & 2.6 & 4.8 & 8.9 & 3.0 & 257.7 & 213.3 & 174.7 \\
TestPromptDC & 8.9 & 9.3 & 2.3 & 3.7 & 9.1 & \underline{2.5} & 185.6 & 153.9 & 138.4 \\

\midrule
\method{}$_\text{LoRA}$ & \textbf{1.8} & \textbf{1.8} & \textbf{1.9} & \textbf{2.4} & \textbf{2.3} & \textbf{2.3} & \textbf{119.4} & \textbf{120.0} & \textbf{118.4} \\
\method{}$_\text{VPT}$ & \textbf{1.8} & \textbf{1.8} & \textbf{1.9} & \textbf{2.4} & \textbf{2.3} & \textbf{2.3} & \underline{119.8} & \textbf{120.0} & \underline{118.6} \\
\bottomrule


\end{tabular}

    }
\vspace{-1em}
\end{table}

\vspace{-0.5em}
\paragraph{Improved temporal consistency.}
Temporal consistency evaluation is shown in~\cref{tab:main_opw} and the qualitative results are presented in~\cref{fig:qualitative_consistency}. \method{} achieves the lowest OPW error in both indoor and outdoor settings, demonstrating the benefit of test-time adaptation. 

\vspace{-0.5em}
\paragraph{Adherence to conditioning vs.~generalization.}
To analyze how different methods respond to the provided 3D conditioning, we separately evaluate the depth predictions at the conditioning points and at all other locations. 
As shown in~\cref{tab:condition_mask}, the errors of most depth completion schemes are significantly lower at the sparse, observed conditioning points, and increase as one moves into the parts that are actually completed.
For example, for OMNI-DC~\cite{zuo2025omni} the increase is nearly 4$\times$ for ScanNet and 2$\times$ for 7-Scenes; indicating overfitting to the provided condition points. With \method{} the performance gap is much smaller---for ScanNet barely noticeable---as test-time adaptation effectively grounds the strong geometric prior of the foundation model and precisely calibrates it using the sparse condition points (\cf first sample in~\cref{fig:qualitative_main}). 

\begin{table}[t]
    \centering
    \caption{\textbf{Comparison of depth errors inside and outside conditioned regions} 
    [AbsRel (\%)$\downarrow$].
    }
    \vspace{-0.5em}
    \label{tab:condition_mask}
    \resizebox{\linewidth}{!}{



\begin{tabular}{l *{6}{c}}
\toprule
& \multicolumn{3}{c}{ScanNet (100)} & \multicolumn{3}{c}{7-Scenes (SfM)} \\
 \cmidrule(lr){2-4} \cmidrule(lr){5-7}
& In Cond. & Out Cond. & Diff. & In Cond. & Out Cond. & Diff. \\

\midrule
PromptDA & 11.8 & 13.2 & 1.4 & 3.0 & 8.3 & 5.3 \\
PriorDA-v1.1 & 1.9 & 2.5 & 0.6 & 2.0 & 3.7 & 1.7 \\
OMNI-DC-v1.1 & \textbf{0.4} & 1.5 & 1.1 & \underline{1.7} & 3.3 & 1.6 \\
Marigold-DC & 2.4 & 3.7 & 1.3 & 1.8 & 4.1 & 2.3 \\
TestPromptDC & 3.1 & 3.8 & 0.7 & \underline{1.7} & 3.1 & 1.4 \\
\midrule
\method{}$_\text{LoRA}$ & \underline{0.8} & \textbf{0.9} & \textbf{0.1} & \textbf{1.6} & \underline{2.8} & \underline{1.2} \\
\method{}$_\text{VPT}$ & 0.9 & \underline{1.0} & \textbf{0.1} & \underline{1.7} & \textbf{2.6} & \textbf{0.9} \\

\bottomrule
\end{tabular}




    }
\end{table}
\vspace{-1em}

\begin{figure}[t]
    \centering
    \includegraphics[width=\linewidth]{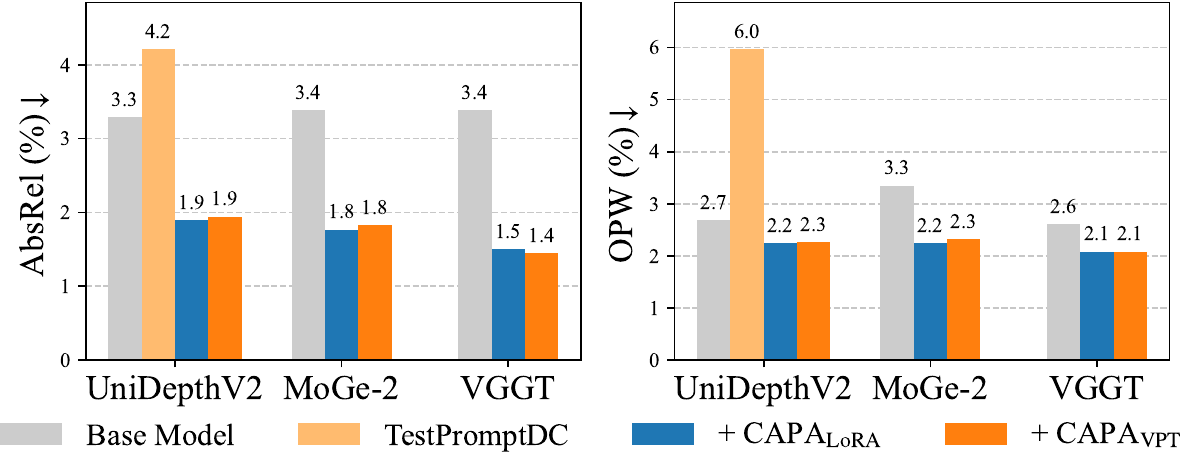}
    \vspace{-1.4em}
    \caption{\textbf{\method{} results when applied to other base models}. 
    }
    \label{fig:other_model}
    \vspace{-1.2em}
\end{figure}

\paragraph{Other base models.}
We apply \method{} to two more base models, UniDepthV2~\cite{piccinelli2025unidepthv2} and MoGe-2~\cite{wang2025moge2}, to demonstrate its versatility. Results, averaged over ScanNet and 7-Scenes, are shown in~\cref{fig:other_model}. For UniDepthV2, we also compare with TestPromptDC~\cite{jeong2025test}, which is also based on prompt tuning and uses the same base model, but injects learnable tokens into each pixel before encoding. This design leads to worse performance, likely due to noise being introduced around the condition points. In contrast, our method consistently improve the base model in terms of both depth accuracy and temporal consistency. Across all three base models, both LoRA and VPT reduce the depth error by $\approx$2$\times$, emphasizing the general validity of our adaptation scheme beyond a specific choice of base model.

\subsection{Ablation Experiments}\label{sec:ablation}
\paragraph{Datasets.}
We randomly sample 32, 8, and 16 sequences from ScanNet, 7-Scenes, and Metropolis, respectively.

\begin{table}[t]
    \centering
    \caption{\textbf{Effect of parameter sharing across video frames}.
    }
    \vspace{-0.5em}
    \label{tab:token_share}
    \resizebox{\linewidth}{!}{

\setlength{\tabcolsep}{2pt} 

\begin{tabular}{cc *{6}{c}}
\toprule
\multirow{2}{*}{} & \multicolumn{1}{c}{\multirow{2}{*}{\makecell{Parameter \\Sharing}}} & \multicolumn{2}{c}{\makecell{ScanNet (100)}}  &\multicolumn{2}{c}{\makecell{7-Scenes (SfM)}} & \multicolumn{2}{c}{\makecell{Metropolis (8-line)}} \\
\cmidrule(lr){3-4} \cmidrule(lr){5-6} \cmidrule(lr){7-8}
& & AbsRel$\downarrow$ & OPW$\downarrow$ & AbsRel$\downarrow$ & OPW$\downarrow$ & AbsRel$\downarrow$ & OPW$\downarrow$ \\

\midrule

                    


\multirow{2}{*}{\makecell{\method{}$_\text{LoRA}$}} 
    & Frame & 1.7 & 4.9 & 4.3 & 4.3 & 8.7 & 153.0 \\
    & Sequence & 0.9 & 1.9 & 3.3 & 2.7 & 7.7 & 117.0 \\
                    
\noalign{\vskip 0.6ex}\cdashline{1-8}\noalign{\vskip 0.6ex}

\multirow{2}{*}{\makecell{\method{}$_\text{VPT}$}}
    & Frame & 1.5 & 4.1 & 3.6 & 4.0 & 8.8 & 149.9 \\
    & Sequence & 1.0 & 1.9 & 2.7 & 2.5 & 8.2 & 119.0 \\
    
\bottomrule
\end{tabular}

    }
\end{table}

\begin{figure}[t]
    \centering
    \includegraphics[width=\linewidth]{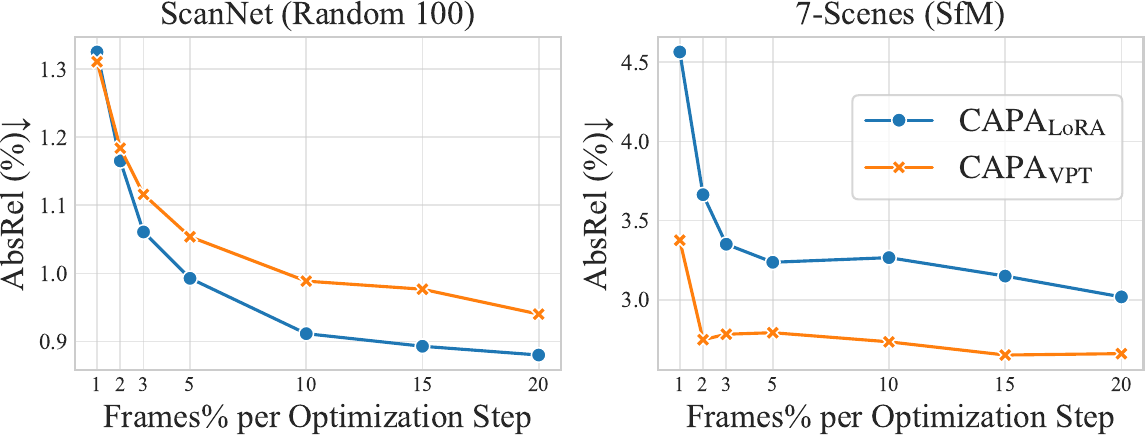}
    \vspace{-1.3em}
    \caption{\textbf{Influence of mini-batch size}. Performance improves with more frames and saturates at $\approx$10\%.
    }
    \label{fig:curve_frame}
\end{figure}

\begin{figure}[!t]
    \centering
    \includegraphics[width=\linewidth]{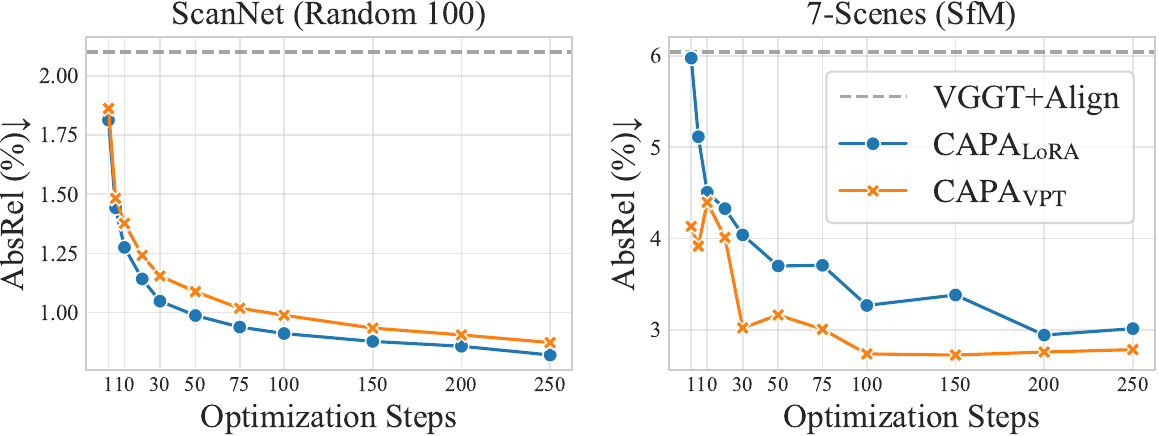}
    \vspace{-1.3em}
    \caption{\textbf{Reconstruction quality vs.\ optimization steps.}
    }
    \label{fig:curve_step}
    \vspace{-1em}
\end{figure}

\vspace{-1em}
\paragraph{Parameter sharing within sequence.}
For video depth completion, our method jointly optimizes over an entire sequence, sharing a single set of learnable parameters across all frames. As shown in~\cref{tab:token_share}, sequence-level adaptation reaches better depth accuracy and temporal consistency than per-frame tuning, particularly when the conditioning is sparse.
We hypothesize that the gain arises because shared parameters encourage the adaptation to aggregate information across frames, and to suppress noise in the 3D condition.
Besides the gain in performance, parameter sharing is also computationally more efficient: model adaptation needs to be done only once per sequence for a fixed number of steps (using mini-batch optimization), then the tuned model can be applied to all frames.
We note that, as more condition points are added, per-frame tuning benefits more from the increased data density, due to the larger effective model capacity. See the appendix for further analysis.

\begin{table}[t]
    \centering
    \caption{\textbf{Batch size vs.\ optimization steps} 
    [AbsRel (\%)$\downarrow$].
    }
    \vspace{-0.5em}
    \label{tab:nstep_nframe}
    \resizebox{\linewidth}{!}{
        \setlength{\tabcolsep}{3pt} 

\begin{tabular}{ccc *{7}{c}}
\toprule

\multirow{2}{*}{} & \multicolumn{1}{c}{\multirow{2}{*}{frames$~[\%$]}} & \multicolumn{1}{c}{\multirow{2}{*}{$\#$~steps}} & \multicolumn{3}{c}{ScanNet} &\multicolumn{2}{c}{7-Scenes} & \multicolumn{2}{c}{Metropolis} \\
\cmidrule(lr){4-6} \cmidrule(lr){7-8} \cmidrule(lr){9-10}

 & \multicolumn{1}{c}{} & \multicolumn{1}{c}{} & \multicolumn{1}{c}{SIFT} & \multicolumn{1}{c}{100} & \multicolumn{1}{c}{\textless{}3m} 
 & \multicolumn{1}{c}{SfM} & \multicolumn{1}{c}{100}
 & \multicolumn{1}{c}{8-line} & \multicolumn{1}{c}{16-line} \\

\midrule

 




 \multirow{3}{*}{\method{}$_\text{LoRA}$}  
 & 40 & 25 & 1.2 & 1.0 & 1.3 & 3.9 & 1.2 & 8.0 & 7.4 \\
 & 20 & 50 & 1.1 & 0.9 & 1.1 & 3.6 & 1.2 & 7.7 & 7.3 \\
 & 10 & 100 & 1.1 & 0.9 & 1.4 & 3.3 & 1.1 & 7.7 & 7.2 \\
\noalign{\vskip 0.6ex}\cdashline{1-10}\noalign{\vskip 0.6ex}
\multirow{3}{*}{\method{}$_\text{VPT}$}  
& 40 & 25 & 1.3 & 1.1 & 1.4 & 3.2 & 1.4 & 8.5 & 8.1 \\
 & 20 & 50 & 1.3 & 1.0 & 1.5 & 3.1 & 1.3 & 8.3 & 7.8 \\
 & 10 & 100 & 1.2 & 1.0 & 1.4 & 2.7 & 1.2 & 8.2 & 7.7 \\

\bottomrule
\end{tabular}
    }
\end{table}

\begin{table}[t]
    \centering
    \caption{\textbf{Comparison to full-model finetuning} [AbsRel (\%)$\downarrow$].
    } 
    \vspace{-0.5em}
    \label{tab:finetune}
    \resizebox{\linewidth}{!}{
        \setlength{\tabcolsep}{4pt} 

\begin{tabular}{c *{8}{c}}
\toprule
 \multicolumn{1}{c}{\multirow{2}{*}{}} & \multirow{2}{*}{param.~$[\%]$} & \multicolumn{3}{c}{ScanNet}  &\multicolumn{2}{c}{7-Scenes} & \multicolumn{2}{c}{Metropolis} \\
\cmidrule(lr){3-5} \cmidrule(lr){6-7} \cmidrule(lr){8-9}
 \multicolumn{2}{c}{} & \multicolumn{1}{c}{SIFT} & \multicolumn{1}{c}{100} & \multicolumn{1}{c}{\textless{}3m} 
 & \multicolumn{1}{c}{SfM}  & \multicolumn{1}{c}{100}
 & \multicolumn{1}{c}{8-line} & \multicolumn{1}{c}{16-line} \\

\midrule

\color{gray} VGGT & \color{gray} - & \color{gray} 2.5 & \color{gray} 2.1 & \color{gray} 2.1 &\color{gray}  6.0 & \color{gray}  5.3 & \color{gray} 10.7 & \color{gray} 10.9 \\
\noalign{\vskip 0.6ex}\cdashline{1-9}\noalign{\vskip 0.6ex}
 FT-All & 100.00 & 1.0 & 0.9 & 1.3 & 3.6 & 1.1 & 7.4 & OOM \\
 FT-Encoder & 32.32 & 1.0 & 0.9 & 1.2 & 3.8 & 1.1 & 7.4 & 6.8 \\
FT-Head & 3.47 & 1.2 & 1.1 & 1.3 & 4.0 & 1.5 & 9.1 & 8.5 \\
\noalign{\vskip 0.6ex}\cdashline{1-9}\noalign{\vskip 0.6ex}
 LoRA-Encoder & 0.04 & 1.1 & 0.9 & 1.4 & 3.3 & 1.1 & 7.7 & 7.2 \\
 VPT-Encoder & 0.04 & 1.2 & 1.0 & 1.4 & 2.7 & 1.2 & 8.2 & 7.7 \\



\bottomrule
\end{tabular}
    }
\end{table}

\paragraph{Mini-batch tuning.}
For video inputs, we randomly sample a subset of frames at each optimization step to form a mini-batch for adapting the shared model. Increasing the number of frames per batch generally boosts performance,  as the model is exposed to a broader range of viewpoints and thus receives more informative and stable gradients; at the cost of additional computation. As shown in~\cref{fig:curve_frame}, using just 2\% of the available frames already delivers a clear improvement over the single-frame (1\%) baseline, while the performance gain gradually saturates beyond $\approx$10\%.

\paragraph{Optimization steps.}
The number of optimization steps determines the quality-cost trade-off. As~\cref{fig:curve_step} shows, a single update step yields noticeable improvement, with performance stabilizing after $\approx$100 steps. Furthermore, analyzing the interaction with mini-batch size (\cref{tab:nstep_nframe}), we conclude that for a fixed budget, more steps (frequent parameter updates) are more effective than larger batches.

\paragraph{Comparison with full model finetuning.}
To establish an upper bound for our test-time adaptation, we also perform full-model finetuning (FT) on each input sequence. We test three variants: finetuning only the encoder (FT-Encoder), only the decoder head (FT-Head), or the entire model (FT-All). As shown in \cref{tab:finetune}, FT-All achieves the highest performance, as expected. However, it only marginally surpasses \method{} while updating $\approx$2500$\times$ more parameters. Interestingly, FT-Head performs worst, despite updating 80$\times$ more parameters than our PEFT approach.

\section{Conclusion}
\label{sec:conclusion}
We have introduced \method{}, a new paradigm where depth completion is reframed as the parameter-efficient adaptation of a 3D foundation model at test time. This model-agnostic approach achieves state-of-the-art performance by grounding strong geometric priors using sparse, scene-specific gradients. While \method{} is significantly more efficient than other test-time optimization baselines, it remains slower than a single forward pass. 
We provide an efficiency analysis in the appendix.
Future work will explore methods like trained optimizers to accelerate convergence.

\clearpage
{
    \small
    \bibliographystyle{ieeenat_fullname}
    \bibliography{main}
}

\clearpage

\appendix
\section*{Appendix}

\renewcommand*{\thesection}{\Alph{section}}
\newcommand{\multiref}[2]{\cref{#1}--\ref{#2}}
\renewcommand{\thetable}{S\arabic{table}}
\renewcommand{\thefigure}{S\arabic{figure}}

\setcounter{table}{0}
\setcounter{figure}{0}

\section{Implementation Details}

\subsection{Hyper-parameters and Practical Settings}
\paragraph{Default settings.} We employ AdamW optimizer with the default setting (\ie $\beta_1=0.9, \beta_2 = 0.999$).
Additionally, we apply gradient clipping with a maximum norm of 1.0 to ensure training stability.
In our default setting, we use 100 optimization steps, using 10\% frames at each step for multi-image input. 
Specifically, for UniDepthV2+\method{}\textsubscript{VPT}, we use 20\% frames, as the improvement from 10\% is noticeable.
The LoRA scaling factor is set to $\alpha = 2 \times r$ across all configurations.
We summarize the LoRA rank, VPT prompt length, and learning rates (lr) for different base models in \cref{tab:hyperparams}. For MoGe-2 and UniDepthV2 with the VPT variant, we apply a linear decay schedule that reduces the learning rate to $0.33\times$ its initial value.

\begin{table}[h]
    \centering
    \caption{\textbf{Hyperparameters for different base models.}}
    \label{tab:hyperparams}
    \resizebox{\columnwidth}{!}{
    \vspace{-0.5em}
    \begin{tabular}{p{3.5cm}p{1.5cm}p{1.5cm}p{1.3cm}p{1.3cm}}
    \toprule
    Base Model & LoRA Rank ($r$) & VPT Length ($t$) & LoRA lr & VPT lr \\
    \midrule
    VGGT~\cite{wang2025vggt} & 4 & 16 & $1\times10^{-3}$ & $2\times10^{-4}$ \\
    MoGe-2~\cite{wang2025moge2} & 16 & 64 & $3\times10^{-4}$ & $2\times10^{-4}$ \\
    UniDepthV2~\cite{piccinelli2025unidepthv2} & 16 & 96 & $3\times10^{-4}$ & $2\times10^{-4}$ \\
    \bottomrule
\end{tabular}
}
\vspace{-1em}
\end{table}

\paragraph{LoRA \vs VPT.}
In ablation studies, we found that both LoRA and VPT achieve comparable accuracy and are robust to hyperparameter changes. 
In practice, VPT performs slightly better when using generic prompts pre-tuned once on a few samples from an auxiliary dataset (\cf \cref{sec:vpt_token_init}). 
These learned prompts serve as a robust initialization that can be applied directly to new scenes, whereas LoRA typically starts with zero-initialized adapters.

\paragraph{Insertion position.} 
We found in practice that the performance is robust to which layers PEFT parameters are inserted into. 
For example, on the ScanNet dataset, applying LoRA to all 24 layers or only the deep 12 layers yields an identical AbsRel error of 0.011. Restricting adaptation to the shallow 12 layers results in a marginal degradation to 0.012. 
We therefore default to inserting parameters into all layers for simplicity and maximizing model capacity.

\subsection{Inference Precision}
We execute the ViT backbone using bfloat16 precision for efficiency. 
However, to maintain the precision of depth predictions, we explicitly cast to float32 for the depth head. 
We find that this mixed-precision approach yields higher accuracy, despite slightly higher inference latency and memory usage.

\begin{table*}[t]
    \centering
    \caption{\textbf{MAE}(\%)$\downarrow$ in quantitative comparison of \method{} with baseline methods.}
    \label{tab:main_mae}
    \resizebox{\linewidth}{!}{
        \setlength{\tabcolsep}{6pt} 

\begin{tabular}{l *{12}{r}}
\toprule
Dataset & \multicolumn{3}{c}{ScanNet} & \multicolumn{3}{c}{7-Scenes} & \multicolumn{3}{c}{iBims} & \multicolumn{3}{c}{Metropolis}  \\
 \cmidrule(lr){1-1} \cmidrule(lr){2-4} \cmidrule(lr){5-7} \cmidrule(lr){8-10} \cmidrule(lr){11-13}
Condition & SIFT & 100 & $<3m$ & SfM & 100 & $<3m$ & SIFT & 100 & $<5m$ & 8-line & 16-line & 32-line \\

\midrule

DepthAnythingV2 & 9.2 & 7.5 & 11.8 & 9.9 & 8.3 & 8.7 & 13.0 & 11.9 & 75.8 & 3569.9 & 3670.8 & 3682.4 \\
UniDepthV2 & 6.1 & 5.3 & 5.9 & 7.6 & 6.7 & 6.7 & 10.6 & 9.1 & 16.0 & 1401.8 & 1316.9 & 1305.3 \\
MoGe-2 & 7.7 & 6.7 & 7.2 & 7.0 & 6.0 & 6.1 & 10.0 & 8.6 & 15.8 & 1307.9 & 1253.7 & 1244.6 \\
VGGT & 4.8 & 4.4 & 4.5 & 8.9 & 8.2 & 8.3 & 14.1 & 12.4 & 20.4 & 861.1 & 839.7 & 833.3 \\
VideoDA & 9.2 & 7.8 & 8.5 & 10.8 & 9.2 & 9.3 & 16.5 & 13.8 & 21.3 & 2341.8 & 2219.1 & 2199.5 \\
\noalign{\vskip 0.6ex}\cdashline{1-13}\noalign{\vskip 0.6ex}
PromptDA & 27.6 & 22.7 & 23.8 & 15.0 & 12.5 & 8.8 & 34.7 & 26.1 & 49.6 & 1597.2 & 1108.9 & 963.8 \\
PriorDA-v1.1 & 5.8 & 4.7 & 26.4 & 6.6 & 5.6 & 21.0 & 9.0 & 8.8 & 72.8 & 1352.5 & 1103.9 & 978.7 \\
OMNI-DC-v1.1 & 3.7 & 3.0 & 14.8 & 6.0 & 4.1 & 11.0 & \underline{5.3} & \textbf{5.3} & 62.1 & 976.7 & 718.1 & 590.9 \\
Marigold-DC & 7.7 & 7.1 & 7.7 & 7.4 & 7.3 & 4.0 & 22.8 & 20.2 & 38.6 & 1389.0 & 945.1 & 710.5 \\
TestPromptDC & 7.7 & 7.1 & 6.5 & 5.7 & 7.2 & 5.2 & 18.4 & 16.1 & 29.1 & 906.8 & 665.9 & 546.5 \\
\midrule
\method{}$_\text{LoRA}$ & \textbf{2.0} & \textbf{1.8} & \underline{3.1} & \underline{5.2} & \textbf{1.9} & \textbf{2.0} & 5.4 & 7.3 & \underline{15.7} & \textbf{456.8} & \textbf{401.5} & \textbf{375.1} \\
\method{}$_\text{LoRA}$ & \underline{2.2} & \underline{1.9} & \textbf{2.7} & \textbf{5.0} & \underline{2.1} & \underline{2.2} & \textbf{4.6} & \underline{6.1} & \textbf{15.4} & \underline{482.3} & \underline{434.2} & \underline{417.4} \\

\bottomrule 

\noalign{\vskip -1.5ex} \\
\noalign{\vskip 0.6ex}\cdashline{1-13}\noalign{\vskip 0.6ex}
\method{}$_\text{LoRA}$ + MoGe-2 & 2.6 & 2.1 & 3.6 & 5.3 & 2.6 & 2.9 & 6.3 & 7.5 & 15.5 & 596.8 & 483.5 & 447.2 \\
\method{}$_\text{VPT}$ + MoGe-2 & 2.9 & 2.4 & 4.0 & 5.4 & 3.1 & 3.5 & 5.0 & 7.0 & 17.5 & 640.1 & 537.4 & 508.1 \\
\noalign{\vskip 0.6ex}\cdashline{1-13}\noalign{\vskip 0.6ex}
\method{}$_\text{LoRA}$ + UniDepthV2 & 2.6 & 2.2 & 3.8 & 5.5 & 2.8 & 3.0 & 7.0 & 8.5 & 19.2 & 594.1 & 480.8 & 447.3 \\
\method{}$_\text{VPT}$ + UniDepthV2 & 2.6 & 2.2 & 4.3 & 5.6 & 2.7 & 3.3 & 6.8 & 8.1 & 21.3 & 611.8 & 510.1 & 475.7 \\
\noalign{\vskip 0.6ex}\cdashline{1-13}\noalign{\vskip 0.6ex}

\end{tabular}

    }
\end{table*}

\begin{table*}[t]
    \centering
    \caption{\textbf{RMSE}(\%)$\downarrow$ in quantitative comparison of \method{} with baseline methods.}
    \label{tab:main_rmse}
    \resizebox{\linewidth}{!}{
        \setlength{\tabcolsep}{6pt} 

\begin{tabular}{l *{12}{r}}
\toprule
Dataset & \multicolumn{3}{c}{ScanNet} & \multicolumn{3}{c}{7-Scenes} & \multicolumn{3}{c}{iBims} & \multicolumn{3}{c}{Metropolis}  \\
 \cmidrule(lr){1-1} \cmidrule(lr){2-4} \cmidrule(lr){5-7} \cmidrule(lr){8-10} \cmidrule(lr){11-13}
Condition & SIFT & 100 & $<3m$ & SfM & 100 & $<3m$ & SIFT & 100 & $<5m$ & 8-line & 16-line & 32-line \\

\midrule

DepthAnythingV2 & 70.1 & 18.8 & 317.6 & 18.4 & 16.0 & 16.2 & 24.4 & 25.0 & 9905.9 & 317854.6 & 363883.5 & 339026.5 \\
UniDepthV2 & 13.0 & 12.3 & 12.9 & 15.6 & 15.0 & 15.0 & 22.6 & 21.8 & \underline{30.2} & 2849.4 & 2845.6 & 2824.4 \\
MoGe-2 & 15.3 & 14.3 & 15.0 & 14.5 & 13.8 & 13.8 & 20.5 & \underline{19.7} & \textbf{28.9} & 2424.5 & 2372.7 & 2370.3 \\
VGGT & 11.6 & 11.3 & 11.5 & 19.6 & 19.2 & 19.3 & 30.2 & 29.7 & 39.5 & 2079.8 & 2045.5 & 2048.1 \\
VideoDA & 15.4 & 14.1 & 15.0 & 18.5 & 16.7 & 16.6 & 29.8 & 28.1 & 37.4 & 3758.6 & 3710.5 & 3667.5 \\
\noalign{\vskip 0.6ex}\cdashline{1-13}\noalign{\vskip 0.6ex}
PromptDA & 36.7 & 30.0 & 33.3 & 24.1 & 19.5 & 16.9 & 51.7 & 42.8 & 82.9 & 2869.2 & 2167.4 & 1943.3 \\
PriorDA-v1.1 & 11.6 & 10.9 & 39.7 & 13.1 & 12.8 & 29.9 & 19.2 & 20.7 & 105.2 & 2370.4 & 2059.7 & 1924.4 \\
OMNI-DC-v1.1 & 9.0 & 8.9 & 31.5 & 12.6 & 11.6 & 24.7 & \textbf{15.3} & \textbf{16.6} & 102.5 & 2002.1 & 1580.7 & 1403.1 \\
Marigold-DC & 16.1 & 17.4 & 18.4 & 14.7 & 16.9 & 11.0 & 44.3 & 48.0 & 70.6 & 2543.1 & 1852.8 & 1406.4 \\
TestPromptDC & 16.0 & 17.2 & 12.3 & 12.0 & 16.8 & 9.7 & 37.7 & 38.7 & 49.9 & 1776.3 & 1357.8 & \textbf{1116.6} \\
\midrule
\method{}$_\text{LoRA}$ & \textbf{5.3} & \textbf{5.3} & \underline{9.8} & 11.9 & \textbf{6.1} & \textbf{6.6} & 18.1 & 25.2 & 34.8 & \textbf{1307.2} & \textbf{1200.4} & \underline{1147.1} \\
\method{}$_\text{LoRA}$ & \underline{5.7} & \underline{5.5} & \textbf{8.9} & \textbf{11.1} & \underline{6.3} & \underline{7.0} & \underline{16.2} & 20.8 & 34.6 & \underline{1348.1} & \underline{1259.2} & 1232.1 \\

\bottomrule 

\noalign{\vskip -1.5ex} \\
\noalign{\vskip 0.6ex}\cdashline{1-13}\noalign{\vskip 0.6ex}
\method{}$_\text{LoRA}$ + MoGe-2 & 6.4 & 6.4 & 10.0 & \underline{11.3} & 7.6 & 8.7 & 17.0 & 23.2 & 31.8 & 1499.6 & 1325.6 & 1260.5 \\
\method{}$_\text{VPT}$ + MoGe-2 & 6.9 & 6.8 & 11.1 & 11.6 & 9.2 & 10.2 & \textbf{15.3} & 22.7 & 37.7 & 1567.1 & 1401.5 & 1356.3 \\
\noalign{\vskip 0.6ex}\cdashline{1-13}\noalign{\vskip 0.6ex}
\method{}$_\text{LoRA}$ + UniDepthV2 & 6.8 & 6.8 & 10.4 & 12.3 & 8.2 & 8.9 & 18.4 & 25.2 & 38.4 & 1494.1 & 1322.1 & 1255.6 \\
\method{}$_\text{VPT}$ + UniDepthV2 & 6.5 & 6.5 & 11.8 & 12.3 & 8.1 & 9.7 & 17.5 & 24.6 & 44.2 & 1529.1 & 1376.2 & 1310.7 \\
\noalign{\vskip 0.6ex}\cdashline{1-13}\noalign{\vskip 0.6ex}

\end{tabular}

    }
\end{table*}

\section{Evaluation Details}
\subsection{Baselines}
We evaluate all baseline methods using their official open-source implementations and public checkpoints. We adhere to their original protocols regarding input resizing and output interpolation.
For VGGT~\cite{wang2025vggt}, we use the predictions from the depth head. 
We employ the \textit{Depth-Anything-V2-Large} checkpoint for DepthAnythingV2~\cite{yang2024depthv2}, \textit{unidepth-v2-vitl14} for UniDepthV2~\cite{piccinelli2025unidepthv2}, and \textit{Metric-Video-Depth-Anything-Large} for VideoDA~\cite{chen2025video}.
Regarding PriorDA~\cite{wang2025depthprior} and OMNI-DC~\cite{zuo2025omni}, we use their \textit{v1.1} version of the codebases and checkpoints as they offer improved performance over the initial releases.
For TestPromptDC~\cite{jeong2025test}, we limit the number of condition points to 75K via random sampling to satisfy GPU memory constraints.

\subsection{Condition Point Selection}
We follow prior work\cite{zuo2025omni, zuo2024ogni, wang2025depthprior} and define the following condition point selection strategies:
\begin{itemize}
    \item \textbf{SfM}: We utilize actual Structure-from-Motion (SfM) points for 7-Scenes dataset, which were pre-computed using COLMAP~\cite{schonberger2016structure, Brachmann2021ICCV} and contain realistic noise. 
    We filter these points by retaining only those below the 75th percentile for the bundle adjustment error.
    To further exclude outliers from reflective surfaces where SfM and sensor measurements are inconsistent, we filter against ground truth using the following consistency thresholds: $\text{AbsRel} < 0.1$, $\text{AbsErr} < 0.1$, and 3D point distance $< 0.1$.
    \item \textbf{SIFT}: As an alternative to SfM, we extract SIFT feature points from each image using OpenCV. These points are then filtered by the ground truth mask to ensure they correspond to valid depth measurements.
    \item \textbf{Random}: We randomly sample 100 pixels from regions with valid ground truth depth.
    \item \textbf{Limited Range}: We select all points with valid depth values within a predefined range.
    \item \textbf{LiDAR}: We simulate LiDAR patterns by uniformly sample $N_L$ (corresponding to $N_L$-line) pitch angles within ground truth range, which is determined by the mean of pitch angle ranges of each image.
    For each sampled angle, we densely sample pixels horizontally to simulate LiDAR scan lines. This resulting mask is intersected with the ground truth mask to ensure that all selected points have valid depth values.
\end{itemize}

To ensure robustness, if the number of selected condition points is fewer than 5, we supplement them by randomly sampling additional points from the ground truth.

\subsection{Noise in Condition}
We follow the noise injection strategy in OMNI-DC\cite{zuo2025omni} to perturb the sampled ground truth depth values.
Specifically, 10\% of the condition points are randomly selected for corruption. 
For this subset, the original depth values are corrupted by adding noise drawn from a uniform distribution bounded by the 10th and 90th percentiles of the image's overall ground truth depth range. 
Note that for 7-Scenes SfM setting, we do not inject this artificial noise.

\begin{table*}[t]
    \centering
    \caption{Quantitative comparison \textbf{under noise-free conditions} (\ie, using ground-truth).
        Numbers are presented in percentage (\%).
}
    \vspace{-0.5em}
    \label{tab:main_no-noise}
    \resizebox{\linewidth}{!}{

\setlength{\tabcolsep}{4pt} 

\begin{tabular}{l *{12}{c}}

\toprule
Dataset & \multicolumn{6}{c}{ScanNet} & \multicolumn{6}{c}{7-Scenes}  \\
\cmidrule(lr){1-1} \cmidrule(lr){2-7} \cmidrule(lr){8-13}
Condition & \multicolumn{2}{c}{SIFT} & \multicolumn{2}{c}{100} & \multicolumn{2}{c}{$<3m$} 
& \multicolumn{2}{c}{\makecell{SfM Masked\\ Ground Truth}} & \multicolumn{2}{c}{100} & \multicolumn{2}{c}{$<3m$} \\
\cmidrule(lr){2-3} \cmidrule(lr){4-5} \cmidrule(lr){6-7} \cmidrule(lr){8-9} \cmidrule(lr){10-11} \cmidrule(lr){12-13}

& AbsRel↓ & MAE↓ & AbsRel↓ & MAE↓ & AbsRel↓ & MAE↓ & AbsRel↓ & MAE↓ & AbsRel↓ & MAE↓ & AbsRel↓ & MAE↓ \\
\midrule

DepthAnythingV2 & 4.9 & 9.5 & 3.8 & 7.5 & 4.9 & 12.7 & 5.2 & 9.6 & 4.6 & 8.3 & 4.6 & 8.7 \\
UniDepthV2 & 3.2 & 6.0 & 2.7 & 5.3 & 2.8 & 5.9 & 4.0 & 7.3 & 3.7 & 6.6 & 3.6 & 6.7 \\
MoGe-2 & 4.0 & 7.6 & 3.3 & 6.6 & 3.3 & 7.2 & 3.6 & 6.5 & 3.3 & 6.0 & 3.3 & 6.0 \\
VGGT & 2.2 & 4.7 & 1.9 & 4.4 & 1.9 & 4.5 & 4.4 & 8.5 & 4.2 & 8.1 & 4.2 & 8.2 \\
VideoDA & 4.9 & 9.1 & 4.1 & 7.7 & 4.2 & 8.5 & 6.0 & 10.7 & 5.1 & 9.1 & 5.0 & 9.2 \\
\noalign{\vskip 0.6ex}\cdashline{1-13}\noalign{\vskip 0.6ex}
PromptDA & 14.9 & 25.5 & 11.1 & 20.1 & 9.7 & 21.2 & 8.0 & 14.7 & 5.8 & 10.7 & 3.2 & 7.6 \\
PriorDA-v1.1 & 2.5 & 4.4 & 1.6 & 3.3 & 9.2 & 24.0 & 3.0 & 5.5 & 2.5 & 4.5 & 8.0 & 18.6 \\
OMNI-DC-v1.1 & 1.8 & 3.4 & 1.4 & 2.8 & 2.7 & 8.5 & 2.4 & 4.5 & 2.1 & 3.8 & 1.9 & 4.8 \\
Marigold-DC & 3.0 & 5.4 & 1.8 & 3.7 & 2.2 & 7.6 & 3.4 & 6.2 & 2.5 & 4.6 & 1.6 & 4.0 \\
TestPromptDC & 1.5 & 2.8 & \underline{1.1} & 2.3 & \textbf{0.9} & \underline{2.9} & 2.0 & 3.9 & 1.7 & 3.3 & \textbf{0.9} & 2.3 \\
\midrule
\method{}$_\text{LoRA}$ & \textbf{1.0} & \textbf{2.0} & \textbf{0.9} & \textbf{1.8} & \underline{1.0} & \underline{2.9} & \textbf{1.3} & \textbf{2.5} & \textbf{1.0} & \textbf{1.8} & \textbf{0.9} & \textbf{2.0} \\
\method{}$_\text{VPT}$ & \underline{1.1} & \underline{2.2} & \textbf{0.9} & \underline{1.9} & \underline{1.0} & \textbf{2.7} & \underline{1.4} & \underline{2.7} & \underline{1.1} & \underline{2.0} & \underline{1.0} & \underline{2.2} \\
\bottomrule
\noalign{\vskip 1.5ex}\cdashline{1-13}\noalign{\vskip 0.6ex}
\method{}$_\text{LoRA}$ + MoGe-2 & 1.3 & 2.5 & 1.0 & 2.1 & 1.3 & 3.4 & 1.9 & 3.7 & 1.3 & 2.5 & 1.3 & 2.8 \\
\method{}$_\text{VPT}$ + MoGe-2 & 1.5 & 2.8 & 1.2 & 2.3 & 1.5 & 3.9 & 2.1 & 4.0 & 1.6 & 3.1 & 1.6 & 3.4 \\
\method{}$_\text{LoRA}$ + UniDepthV2 & 1.4 & 2.5 & 1.1 & 2.2 & 1.3 & 3.7 & 2.0 & 3.8 & 1.4 & 2.6 & 1.3 & 2.8 \\
\method{}$_\text{VPT}$ + UniDepthV2 & 1.4 & 2.5 & 1.1 & 2.1 & 1.4 & 4.1 & 2.1 & 4.1 & 1.4 & 2.6 & 1.5 & 3.2 \\
\noalign{\vskip 0.6ex}\cdashline{1-13}\noalign{\vskip 0.6ex}

\noalign{\vskip 1.5ex} \\

\toprule
Dataset & \multicolumn{6}{c}{iBims} & \multicolumn{6}{c}{Metropolis}  \\
\cmidrule(lr){1-1} \cmidrule(lr){2-7} \cmidrule(lr){8-13}
Condition & \multicolumn{2}{c}{SIFT} & \multicolumn{2}{c}{100} & \multicolumn{2}{c}{$<5m$} 
& \multicolumn{2}{c}{8-line} & \multicolumn{2}{c}{16-line} & \multicolumn{2}{c}{32-line} \\
\cmidrule(lr){2-3} \cmidrule(lr){4-5} \cmidrule(lr){6-7} \cmidrule(lr){8-9} \cmidrule(lr){10-11} \cmidrule(lr){12-13}
& AbsRel↓ & MAE↓ & AbsRel↓ & MAE↓ & AbsRel↓ & MAE↓ & AbsRel↓ & MAE↓ & AbsRel↓ & MAE↓ & AbsRel↓ & MAE↓ \\
\midrule

DepthAnythingV2 & 4.0 & 13.1 & 3.3 & 12.1 & 9.8 & 75.5 & 59.4 & 3964.5 & 56.7 & 3794.1 & 60.0 & 4018.0 \\
UniDepthV2 & 3.4 & 10.7 & 2.7 & 9.1 & 3.2 & 16.1 & 29.1 & 1385.3 & 28.7 & 1315.0 & 28.0 & 1301.8 \\
MoGe-2 & 3.3 & 9.9 & 2.6 & 8.5 & 3.1 & 15.9 & 21.5 & 1302.3 & 20.1 & 1250.4 & 19.8 & 1242.9 \\
VGGT & 4.0 & 14.1 & 3.3 & 12.3 & 3.8 & 20.3 & 10.1 & 858.2 & 10.2 & 839.4 & 9.9 & 832.5 \\
VideoDA & 5.0 & 16.0 & 3.9 & 13.6 & 4.5 & 20.9 & 49.2 & 2336.6 & 45.9 & 2221.2 & 44.9 & 2199.2 \\
\noalign{\vskip 0.6ex}\cdashline{1-13}\noalign{\vskip 0.6ex}
PromptDA & 9.5 & 29.3 & 5.6 & 20.2 & 6.8 & 46.7 & 24.3 & 1597.2 & 16.1 & 1128.3 & 13.7 & 1014.5 \\
PriorDA-v1.1 & 2.1 & 6.8 & 1.7 & 6.0 & 13.5 & 67.5 & 21.1 & 1128.9 & 14.1 & 797.5 & 10.1 & 604.6 \\
OMNI-DC-v1.1 & 1.5 & 4.9 & \underline{1.4} & 4.9 & 5.8 & 40.6 & 13.5 & 890.3 & 10.0 & 628.2 & 7.5 & 470.0 \\
Marigold-DC & 4.9 & 14.3 & 2.5 & 8.8 & 5.1 & 38.3 & 27.9 & 1267.6 & 18.8 & 807.5 & 13.7 & 571.2 \\
TestPromptDC & 2.1 & 6.6 & 1.6 & 5.7 & \underline{2.5} & 19.9 & 13.0 & 743.9 & 8.9 & 481.9 & \underline{6.8} & \textbf{354.5} \\
\midrule
\method{}$_\text{LoRA}$& \textbf{1.2} & \textbf{4.1} & \textbf{1.3} & \underline{4.7} & \textbf{1.8} & \underline{14.4} & \textbf{7.7} & \textbf{454.1} & \textbf{7.0} & \textbf{398.9} & \textbf{6.7} & \underline{373.8} \\
 \method{}$_\text{VPT}$ & \underline{1.3} & \underline{4.3} & \textbf{1.3} & \textbf{4.6} & \textbf{1.8} & \textbf{14.0} & \underline{8.1} & \underline{478.7} & \underline{7.6} & \underline{432.8} & 7.4 & 416.3 \\
\bottomrule

\noalign{\vskip 1.5ex}\cdashline{1-13}\noalign{\vskip 0.6ex}
\method{}$_\text{LoRA}$ + MoGe-2 &  1.2 & 4.0 & 1.2 & 4.2 & 1.9 & 14.2 & 10.0 & 601.6 & 8.4 & 482.5 & 7.9 & 445.7 \\
\method{}$_\text{VPT}$ + MoGe-2 & 1.3 & 4.2 & 1.2 & 4.5 & 2.1 & 15.9 & 10.9 & 638.6 & 9.6 & 533.9 & 9.1 & 505.9 \\
\method{}$_\text{LoRA}$ + UniDepthV2 & 1.3 & 4.4 & 1.3 & 4.7 & 2.6 & 19.6 & 9.8 & 585.8 & 8.2 & 481.6 & 7.9 & 447.0 \\
\method{}$_\text{VPT}$ + UniDepthV2 & 1.5 & 4.7 & 1.3 & 4.7 & 2.7 & 21.2 & 10.1 & 605.8 & 8.8 & 514.1 & 8.3 & 473.5 \\
\noalign{\vskip 0.6ex}\cdashline{1-13}\noalign{\vskip 0.6ex}

\end{tabular}

    }
\end{table*}

\begin{table*}[h]
    \centering
    \caption{\textbf{Conditioned on various densities}. Numbers are presented in percentage (\%).
    }
    \label{tab:density}
    \resizebox{\linewidth}{!}{
        \setlength{\tabcolsep}{4pt} 

\begin{tabular}{l *{14}{c}}
\toprule

Num. Sampled Points & \multicolumn{2}{c}{50} & \multicolumn{2}{c}{100} & \multicolumn{2}{c}{500} & \multicolumn{2}{c}{1000} & \multicolumn{2}{c}{1500} & \multicolumn{2}{c}{2000} & \multicolumn{2}{c}{2500} \\
\cmidrule(lr){2-3} \cmidrule(lr){4-5} \cmidrule(lr){6-7} \cmidrule(lr){8-9} \cmidrule(lr){10-11} \cmidrule(lr){12-13} \cmidrule(lr){14-15}  
& AbsRel$\downarrow$ & OPW$\downarrow$ & AbsRel$\downarrow$ & OPW$\downarrow$ & AbsRel$\downarrow$ & OPW$\downarrow$ & AbsRel$\downarrow$ & OPW$\downarrow$ & AbsRel$\downarrow$ & OPW$\downarrow$ & AbsRel$\downarrow$ & OPW$\downarrow$ & AbsRel$\downarrow$ & OPW$\downarrow$ \\

\midrule




VGGT & 2.1 & \underline{2.6} & 2.1 & \underline{2.4} & 2.1 & \underline{2.3} & 2.1 & \underline{2.3} & 2.1 & 2.2 & 2.1 & 2.2 & 2.1 & 2.2 \\
PriorDA-v1.1 & 3.1 & 6.0 & 2.5 & 4.9 & 1.8 & 3.5 & 1.7 & 3.3 & 1.7 & 3.1 & 1.7 & 3.1 & 1.7 & 3.1 \\
OMNI-DC-v1.1 & 1.9 & 3.7 & 1.5 & 3.1 & 1.0 & 2.5 & \underline{0.8} & \underline{2.3} & \underline{0.8} & 2.3 & \textbf{0.7} & 2.3 & \underline{0.7} & 2.2 \\
Marigold-DC & 5.0 & 12.4 & 3.7 & 9.7 & 1.8 & 4.6 & 1.4 & 3.6 & 1.3 & 3.2 & 1.2 & 3.1 & 1.1 & 2.9 \\
TestPromptDC & 4.4 & 11.1 & 4.0 & 10.0 & 3.1 & 6.3 & 2.9 & 5.0 & 2.8 & 4.3 & 2.8 & 3.9 & 2.7 & 3.6 \\
\midrule
\method{}$_\text{LoRA}$ (per-image)  & 2.4 & 6.5 & 1.7 & 4.9 & \underline{0.9} & 2.8 & \textbf{0.7} & 2.4 & \textbf{0.7} & 2.3 & \textbf{0.7} & 2.2 & \textbf{0.6} & 2.2 \\
\method{}$_\text{LoRA}$ (per-sequence) & \textbf{1.0} & \textbf{2.0} & \textbf{0.9} & \textbf{1.9} & \textbf{0.8} & \textbf{1.9} & \underline{0.8} & \textbf{1.9} & \underline{0.8} & \textbf{1.8} & \underline{0.8} & \textbf{1.8} & 0.8 & \textbf{1.8} \\
\method{}$_\text{VPT}$ (per-image)  & 2.0 & 5.5 & 1.5 & 4.1 & \underline{0.9} & 2.6 & \underline{0.8} & \underline{2.3} & \textbf{0.7} & 2.2 & \textbf{0.7} & 2.1 & \underline{0.7} & 2.1 \\
\method{}$_\text{VPT}$ (per-sequence)  & \underline{1.1} & \textbf{2.0} & \underline{1.0} & \textbf{1.9} & \underline{0.9} & \textbf{1.9} & 0.9 & \textbf{1.9} & 0.9 & \underline{1.9} & 0.9 & \underline{1.9} & 0.9 & \underline{1.9} \\

\bottomrule 
\end{tabular}

    }
\end{table*}

\begin{table}[h]
    \centering
    \caption{Temporal consistency evaluation [OPW\%$\downarrow$] \textbf{under noise-free conditions}.
    }
    \vspace{-0.5em}
    \label{tab:main_opw_no-noise}
    \resizebox{\linewidth}{!}{
        \setlength{\tabcolsep}{3pt} 

\begin{tabular}{l *{12}{c}}
\toprule
Dataset & \multicolumn{3}{c}{ScanNet} & \multicolumn{3}{c}{7-Scenes} & \multicolumn{3}{c}{Metropolis}  \\
 \cmidrule(lr){1-1} \cmidrule(lr){2-4} \cmidrule(lr){5-7} \cmidrule(lr){8-10} 
Condition & SIFT & 100 & $<3m$ & SfM & 100 & $<3m$ & 8-line & 16-line & 32-line \\

\midrule


UniDepthV2 & 2.7 & 2.5 & 2.5 & \underline{2.9} & \underline{2.9} & 2.6 & 165.2 & \underline{152.2} & 141.1 \\
MoGe-2 & 4.1 & 3.7 & 3.6 & 3.1 & \underline{2.9} & 2.7 & 219.3 & 205.3 & 199.1 \\
VGGT & \underline{2.3} & \underline{2.2} & \underline{2.0} & 3.1 & 3.2 & 2.9 & 149.4 & 152.5 & 149.0 \\
VideoDA & 3.3 & 3.2 & 2.8 & 3.3 & 3.6 & 2.9 & 179.3 & 160.4 & 141.5 \\
\noalign{\vskip 0.6ex}\cdashline{1-13}\noalign{\vskip 0.6ex}
PriorDA-v1.1 & 5.0 & 4.7 & 3.7 & 3.5 & 4.9 & 4.1 & 268.9 & 233.2 & 195.0 \\
OMNI-DC-v1.1 & 3.2 & 3.0 & 9.2 & 3.3 & 3.7 & 10.0 & 171.2 & 174.5 & 171.3 \\
Marigold-DC & 8.1 & 9.0 & 2.6 & 4.8 & 8.9 & 3.0 & 257.7 & 213.3 & 174.7 \\
TestPromptDC & 8.9 & 9.3 & 2.3 & 3.7 & 9.1 & \underline{2.5} & 185.6 & 153.9 & 138.4 \\
\midrule
\method{}$_\text{LoRA}$ & \textbf{1.8} & \textbf{1.8} & \textbf{1.9} & \textbf{2.4} & \textbf{2.3} & \textbf{2.3} & \textbf{119.4} & \textbf{120.0} & \textbf{118.4} \\
\method{}$_\text{VPT}$ & \textbf{1.8} & \textbf{1.8} & \textbf{1.9} & \textbf{2.4} & \textbf{2.3} & \textbf{2.3} & \underline{119.8} & \textbf{120.0} & \underline{118.6} \\
\bottomrule 

\noalign{\vskip 1.5ex}\cdashline{1-13}\noalign{\vskip 0.6ex}
\method{}$_\text{LoRA}$ + MoGe-2 & 2.1 & 2.0 & 2.0 & 2.5 & 2.5 & 2.4 & 120.9 & 118.3 & 116.3 \\
\method{}$_\text{VPT}$ + MoGe-2 & 2.2 & 2.1 & 2.1 & 2.5 & 2.5 & 2.5 & 121.9 & 119.9 & 118.4 \\
\noalign{\vskip 0.6ex}\cdashline{1-13}\noalign{\vskip 0.6ex}

\method{}$_\text{LoRA}$ + UniDepthV2 & 2.1 & 2.0 & 2.0 & 2.5 & 2.5 & 2.4 & 119.8 & 120.9 & 119.3 \\
\method{}$_\text{VPT}$ + UniDepthV2 & 2.1 & 2.0 & 2.0 & 2.5 & 2.6 & 2.4 & 122.6 & 119.9 & 118.9 \\
\noalign{\vskip 0.6ex}\cdashline{1-13}\noalign{\vskip 0.6ex}

\end{tabular}

    }
\end{table}

\section{Experimental Results}
\subsection{Mean Absolute Error}
In \cref{tab:main_mae} and \cref{tab:main_rmse}, we report MAE and RMSE for the quantitative comparison, as a supplement to AbsRel reported in the main table.

\vspace{-1em}

\subsection{Results under Noise-free Conditions}
We also evaluate and compare the results under noise-free conditions (\ie utilizing exact ground truth values as conditions).
Accuracy metrics (AbsRel and MAE) are reported in \cref{tab:main_no-noise}, and temporal consistency metric OPW is reported in \cref{tab:main_opw_no-noise}.
\vspace{-0.3em}

\subsection{Varying condition point density}
\vspace{-0.2em}
While the main paper focuses on sparse or incomplete settings, we extend our analysis here to varying point densities. 
Specifically, we evaluate performance on the ablation subset of ScanNet by randomly sampling condition points at various densities, with noise injection applied.

\begin{figure}[h]
    \centering
    \includegraphics[width=\linewidth]{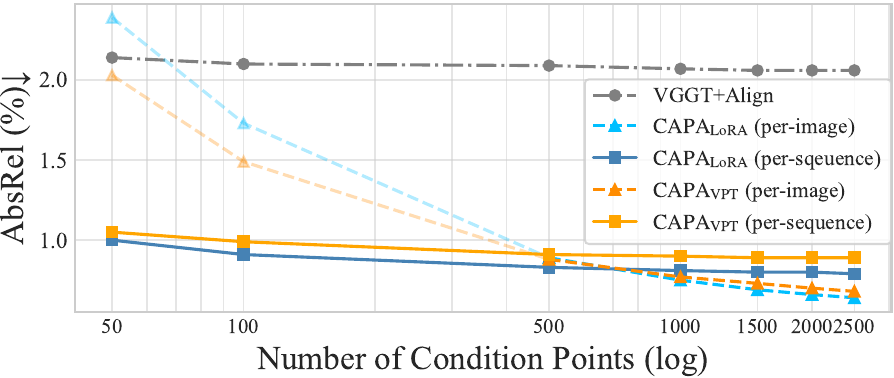}
    \caption{\textbf{Per-frame tuning vs. per-sequence tuning} under various condition point density.
    When the condition is not very sparse ($>500$), per-image tuning is preferred.
    }
    \label{fig:curve_density}
\end{figure}

From \cref{fig:curve_density} and \cref{tab:density} we can see that given denser condition points ($>500$) points, per-image optimization benefit more in terms of accuracy, while per-sequence tuning (with shared parameters) still maintain the optimal temporal consistency.

\subsection{Impact of LoRA Rank and VPT Token Length}
We study the impact of the rank of LoRA and the token length of VPT, the results are reported in \cref{tab:rank_tokenlen}. \method{} is robust to parameter changes, maintaining performance with as few as 0.2M trainable parameters ($r=2$ or $t=8$).

\begin{table}[h!]
    \centering
    \caption{\textbf{Ablation of LoRA rank and VPT token length }[AbsRel\%$\downarrow$].
    }
    \label{tab:rank_tokenlen}
    \resizebox{\linewidth}{!}{
        \setlength{\tabcolsep}{4pt} 

\begin{tabular}{c *{9}{c}}
\toprule
 \multicolumn{1}{c}{\multirow{2}{*}{Method}} & \multirow{2}{*}{\makecell{Rank /\\Token Len.}} & \multirow{2}{*}{\#Param.}& \multicolumn{3}{c}{ScanNet}  &\multicolumn{2}{c}{7-Scenes} & \multicolumn{2}{c}{Metropolis} \\
\cmidrule(lr){4-6} \cmidrule(lr){7-8} \cmidrule(lr){9-10}
 \multicolumn{3}{c}{} & \multicolumn{1}{c}{SIFT} & \multicolumn{1}{c}{100} & \multicolumn{1}{c}{\textless{}3m} 
 & \multicolumn{1}{c}{SfM}  & \multicolumn{1}{c}{100}
 & \multicolumn{1}{c}{8-line} & \multicolumn{1}{c}{16-line} \\

\midrule


\multirow{4}{*}{\method{}$_\textbf{LoRA}$} 
& $r=2$ & 0.20M & 1.1\color{gray}0 & 0.9\color{gray}2 & 1.4\color{gray}2 & 3.8\color{gray}8 & 1.1\color{gray}1 & 7.8\color{gray}4 & 7.2\color{gray}4 \\
& $r=4$ & 0.39M & 1.0\color{gray}8 & 0.9\color{gray}1 & 1.4\color{gray}5 & 3.2\color{gray}7 & 1.1\color{gray}2 & 7.7\color{gray}2 & 7.1\color{gray}7 \\
& $r=8$ & 0.79M & 1.0\color{gray}7 & 0.9\color{gray}0 & 1.2\color{gray}2 & 3.5\color{gray}2 & 1.1\color{gray}1 & 7.7\color{gray}6 & 7.0\color{gray}2 \\
& $r=16$ & 1.57M & 1.1\color{gray}0 & 0.9\color{gray}5 & 1.3\color{gray}8 & 2.9\color{gray}0 & 1.0\color{gray}9 & 7.8\color{gray}4 & 7.0\color{gray}0 \\
\noalign{\vskip 0.6ex}\cdashline{1-9}\noalign{\vskip 0.6ex}
\multirow{4}{*}{\method{}$_\textbf{VPT}$} 
& $t=8$ & 0.20M & 1.2\color{gray}1 & 1.0\color{gray}0 & 1.4\color{gray}7 & 3.0\color{gray}7 & 1.2\color{gray}4 & 8.4\color{gray}6 & 7.8\color{gray}5 \\
& $t=16$ & 0.39M & 1.1\color{gray}8 & 0.9\color{gray}8 & 1.4\color{gray}6 & 3.3\color{gray}5 & 1.2\color{gray}6 & 8.5\color{gray}9 & 7.9\color{gray}9 \\
& $t=32$ & 0.79M & 1.1\color{gray}8 & 1.0\color{gray}2 & 1.4\color{gray}2 & 2.8\color{gray}0 & 1.2\color{gray}7 & 11.1\color{gray}6 & 7.9\color{gray}8 \\
& $t=64$ & 1.57M & 1.3\color{gray}2 & 1.0\color{gray}7 & 1.4\color{gray}6 & 4.0\color{gray}7 & 1.3\color{gray}1 & 11.4\color{gray}0 & 9.4\color{gray}4 \\

\bottomrule
\end{tabular}
    }
\end{table}

\subsection{VPT Token Initialization.} \label{sec:vpt_token_init}
We study different variants of token initialization for Visual Prompt Tuning. The results are reported in \cref{tab:token_init}.
We find that with random initialization like Xavier~\cite{glorot2010understanding}, \method{} already gives significant improvement over the base model. When initialized with pre-tuned tokens, the performance can be further boosted in most cases. Here we pre-tune these tokens on just a few ScanNet++~\cite{yeshwanthliu2023scannetpp} sequences. 

\begin{table}[h!]
    \centering
    \caption{\textbf{Ablation of VPT token initialization}.
    }
    \vspace{-0.5em}
    \label{tab:token_init}
    \resizebox{\linewidth}{!}{
        \setlength{\tabcolsep}{4pt} 

\begin{tabular}{cc *{7}{c}}
\toprule
& \multicolumn{1}{c}{\multirow{2}{*}{\makecell{Token\\Initialization}}} & \multicolumn{3}{c}{ScanNet}  &\multicolumn{2}{c}{7-Scenes} & \multicolumn{2}{c}{Metropolis} \\
\cmidrule(lr){3-5} \cmidrule(lr){6-7} \cmidrule(lr){8-9}
 & \multicolumn{1}{c}{} & \multicolumn{1}{c}{SIFT} & \multicolumn{1}{c}{100} & \multicolumn{1}{c}{\textless{}3m} 
 & \multicolumn{1}{c}{SfM}  & \multicolumn{1}{c}{100}
 & \multicolumn{1}{c}{8-line} & \multicolumn{1}{c}{16-line} \\

\midrule

 


& {\color{gray} Original + Align} & {\color{gray} 2.5} & {\color{gray} 2.1} & {\color{gray} 2.1} & {\color{gray} 6.0} & {\color{gray} 5.3} & {\color{gray} 10.7} & {\color{gray} 10.9} \\
& Xavier & 1.2 & 1.0 & 1.5 & 3.3 & 1.3 & 8.6 & 8.0 \\
& Pre-tuned & 1.2 & 1.0 & 1.4 & 2.7 & 1.2 & 8.2 & 7.7 \\


\bottomrule
\end{tabular}
    }
\end{table}

\subsection{Parameter Sharing for MoGe-2}
We further validate the generalizability of our parameter sharing method by applying it to MoGe-2~\cite{wang2025moge2}, a single-image depth estimation model. 
The results in \cref{tab:token_share_moge} demonstrate that this technique is not limited to multi-view architectures; it significantly improves both the accuracy and temporal consistency of the single-image base model.

\begin{table}[h!]
    \centering
    \caption{\textbf{Effect of parameter sharing across video frames}, taking MoGe-2 as base model.
    }
    \vspace{-0.5em}
    \label{tab:token_share_moge}
    \resizebox{\linewidth}{!}{
        
\setlength{\tabcolsep}{2pt} 

\begin{tabular}{cc *{6}{c}}
\toprule
\multirow{2}{*}{Method} & \multicolumn{1}{c}{\multirow{2}{*}{\makecell{Parameter \\Sharing}}} & \multicolumn{2}{c}{\makecell{ScanNet (100)}}  &\multicolumn{2}{c}{\makecell{7-Scenes (SfM)}} & \multicolumn{2}{c}{\makecell{Metropolis (8-line)}} \\
\cmidrule(lr){3-4} \cmidrule(lr){5-6} \cmidrule(lr){7-8}
& & AbsRel$\downarrow$ & OPW$\downarrow$ & AbsRel$\downarrow$ & OPW$\downarrow$ & AbsRel$\downarrow$ & OPW$\downarrow$ \\

\midrule

\multirow{2}{*}{\makecell{MoGe-2 + \\\method{}$_\text{LoRA}$}} 
    & Frame & 1.9 & 5.0 & 3.5 & 3.6 & 11.0 & 164.7 \\
    & Sequence & 1.1 & 2.1 & 3.1 & 2.6 & 9.9 & 118.2 \\
                    
\noalign{\vskip 0.6ex}\cdashline{1-8}\noalign{\vskip 0.6ex}

\multirow{2}{*}{\makecell{MoGe-2 + \\\method{}$_\text{VPT}$}}
    & Frame & 1.8 & 4.8 & 3.5 & 3.7 & 11.4 & 168.4 \\
    & Sequence & 1.2 & 2.2 & 3.2 & 2.7 & 11.0 & 118.6 \\
    
\bottomrule
\end{tabular}

    }

\end{table}

\subsection{Runtime and Efficiency Analysis}

\paragraph{Inference latency.}
In \cref{tab:runtime}, we analyze the total wall-clock time required to process a 100-frame video sequence from ScanNet. 
While test-time adaptation (TTA) inherently introduces an optimization overhead compared to a single forward pass of the base model, \method{} is significantly more efficient ($> 10\times$) than state-of-the-art TTA competitors like Marigold-DC and TestPromptDC.

\paragraph{Accuracy-cost trade-off.}
\Cref{fig:curve_step} demonstrates the controllability of \method{}'s computational cost. 
A few optimization steps already yield noticeable improvements over the baseline. Performance stabilizes after approximately 100 steps. This allows users to trade off accuracy for speed based on their specific latency constraints, a flexibility not offered by fixed-cost diffusion models.

\paragraph{Memory efficiency.}
Unlike full model fine-tuning (FT-All), which requires storing gradients and optimizer states for the entire backbone, \method{} minimizes the memory (VRAM) footprint during optimization by adopting PEFT techniques, which update only 0.39M parameters ($\approx$0.04\% of the model), reducing peak VRAM usage. 
This allows for the adaptation of large-scale foundation models on standard consumer-grade GPUs, where full fine-tuning may result in Out-Of-Memory (OOM).

\begin{figure}[h!]
    \centering
    \captionof{table}{\textbf{Runtime} on a 100-frame ScanNet video [sec.]}
    \label{tab:runtime}
    \resizebox{\linewidth}{!}{

\begin{tabular}{l ccc}
\toprule
Method & Optimization time & Inference Time & Total Time \\
\midrule
\color{gray}VGGT~\cite{wang2025vggt} & \color{gray}- & \color{gray}9 & \color{gray}9 \\
Marigold-DC~\cite{viola2024marigold} & 1615 & - & 1615 \\
TestPromptDC~\cite{jeong2025test} &  2631 & 40 & 2671 \\
\method{}\textsubscript{LoRA} & 140  & 9 & 149\\
\method{}\textsubscript{VPT} & 146 & 9 & 155 \\
\bottomrule
\end{tabular}
    }
\end{figure}

\subsection{Application to Generative Novel View Synthesis}
\begin{figure}[h]
    \centering
    \includegraphics[width=\linewidth]{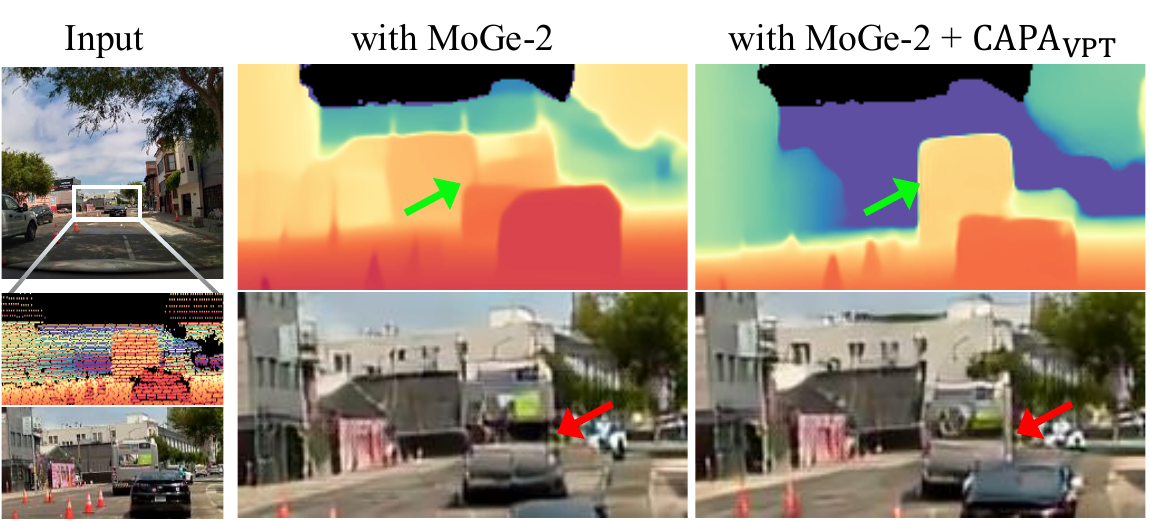}
    \caption{\textbf{Novel view synthesis results} using Gen3C, with depth map from Moge-2 and \method{}$_\text{VPT}$. The viewpoint is moved by 1m higher than the input.}
    \label{fig:gen3c}
\end{figure}

We demonstrate the practical impact of our refined depth maps on downstream novel view synthesis using the Gen3C model~\cite{ren2025gen3c}, as illustrated in~\cref{fig:gen3c}. 
By default, Gen3C relies on standard MoGe-2 depth estimates (indicated by \textcolor{green}{green} arrows), which may produce noticeable synthesis artifacts (indicated by \textcolor{red}{red} arrows). 
Replacing this input with our depth predictions, conditioned on sparse LiDAR, effectively corrects the geometry, resulting in synthesized views that are geometrically consistent and largely free of artifacts.

\subsection{Failure Cases}
\begin{figure}[h]
    \centering
    \includegraphics[width=\linewidth]{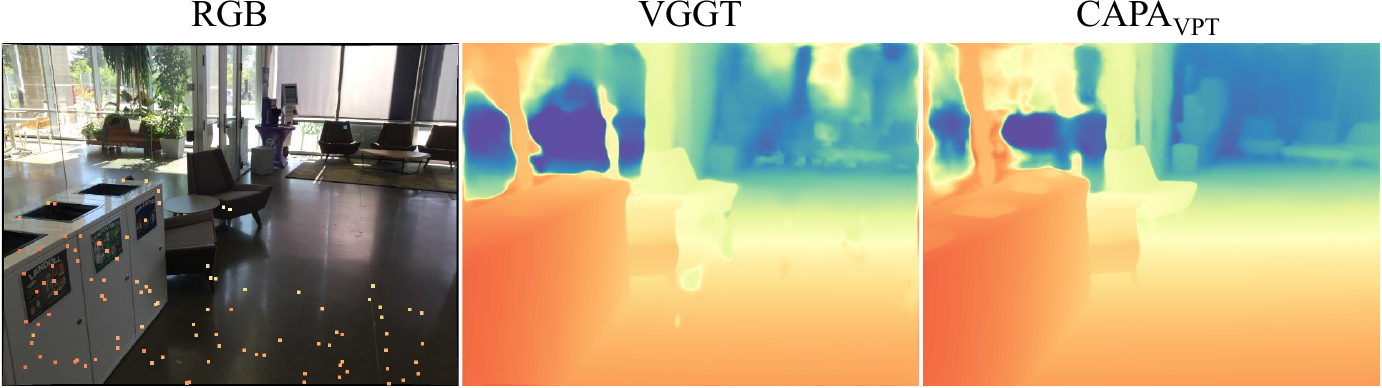}
    \caption{\textbf{Failure case,} where the base model fails on mirrors and \method{} can not fully recover due to insufficient condition points.}
    \label{fig:failure}
\end{figure}

We note that \method{} relies on the geometric prior of the pre-trained base model. Consequently, in extreme scenarios where the foundation model fails completely, such as on visual illusions or complex non-Lambertian reflective surfaces (\eg in \cref{fig:failure}), the adaptation may be insufficient to fully correct the geometry using only sparse cues, potentially resulting in noisy or blurry predictions. 
Furthermore, challenges remain when the physical definition of depth differs between the visual prior (typically surface depth) and the sparse measurements (\eg, LiDAR penetrating glass or measuring reflected distance).

\subsection{Additional Qualitative Results}
\begin{figure*}[t]
    \vspace{-1em}
    \centering
    \includegraphics[width=0.99\linewidth]{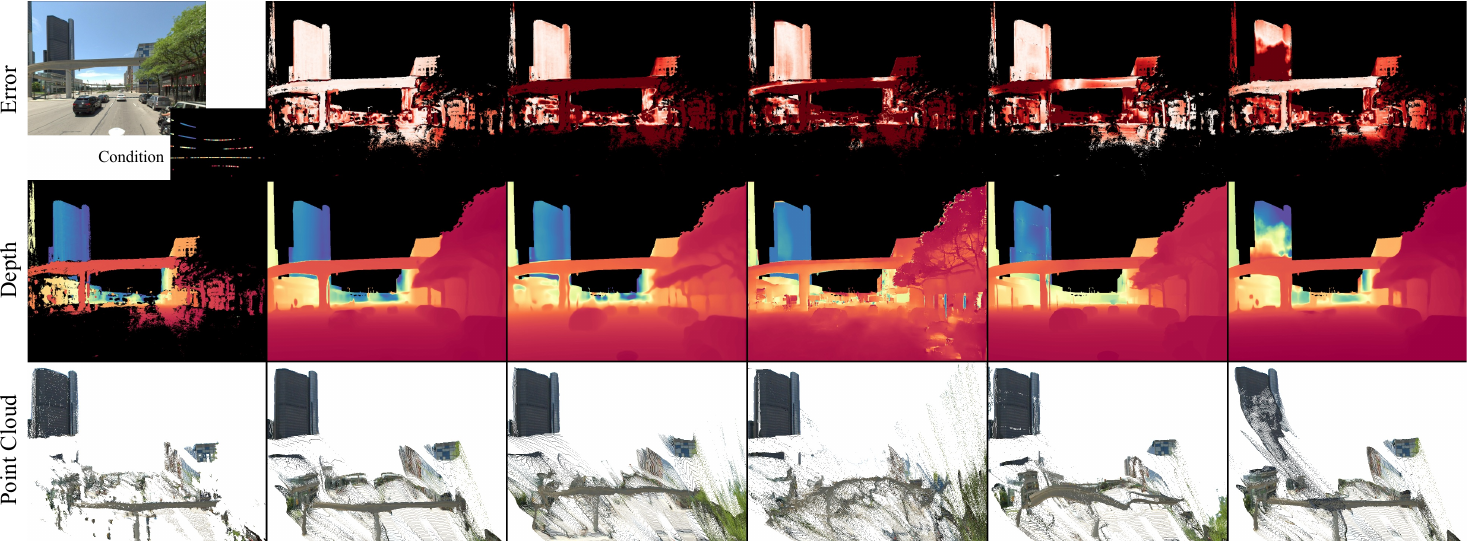}
    \vspace{0.1em} \hrule \vspace{0.1em}
    \includegraphics[width=0.99\linewidth]{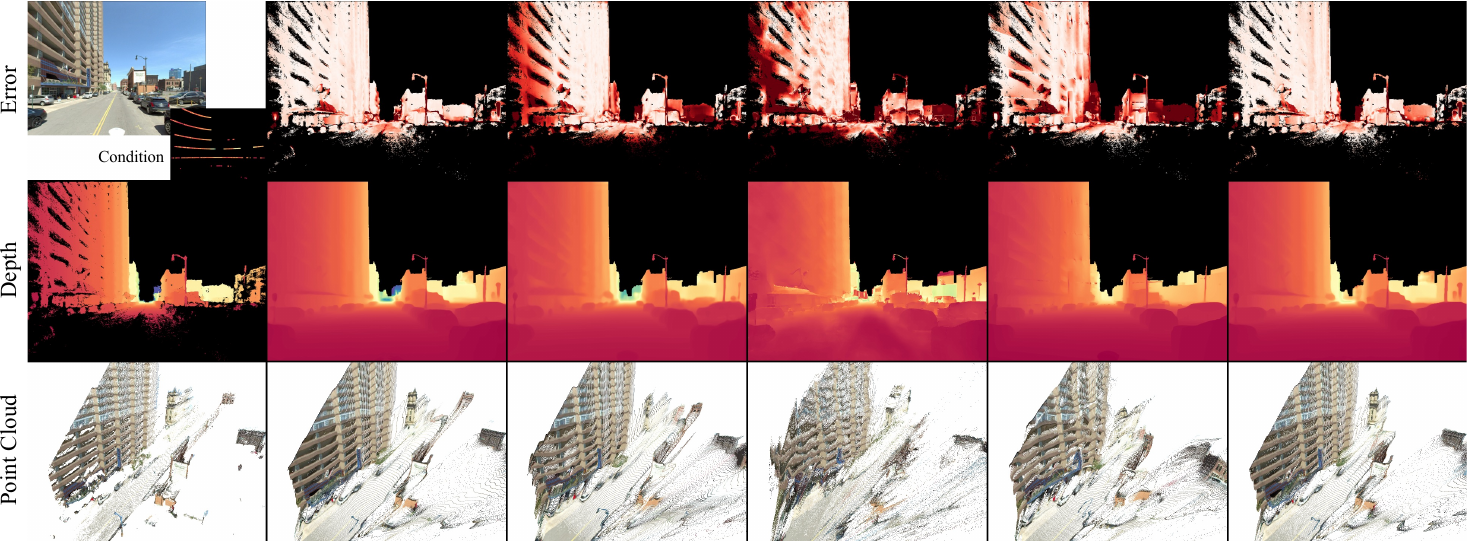}
\caption{Additional \textbf{qualitative comparison on iBims, ScanNet, 7-Scenes, and Metropolis datasets}. Depth is color-coded near~\depth~far, errors low~\error~high.}
  \label{fig:qualitative_supp}
\end{figure*}

Additional qualitative comparisons on the test datasets are shown in \cref{fig:qualitative_supp}, including error maps, depth maps, and point clouds.
Interactive comparisons can be found on the attached website.

\clearpage

\begin{figure*}[h]
    \vspace{-1em}
    \centering
    \includegraphics[width=0.99\linewidth]{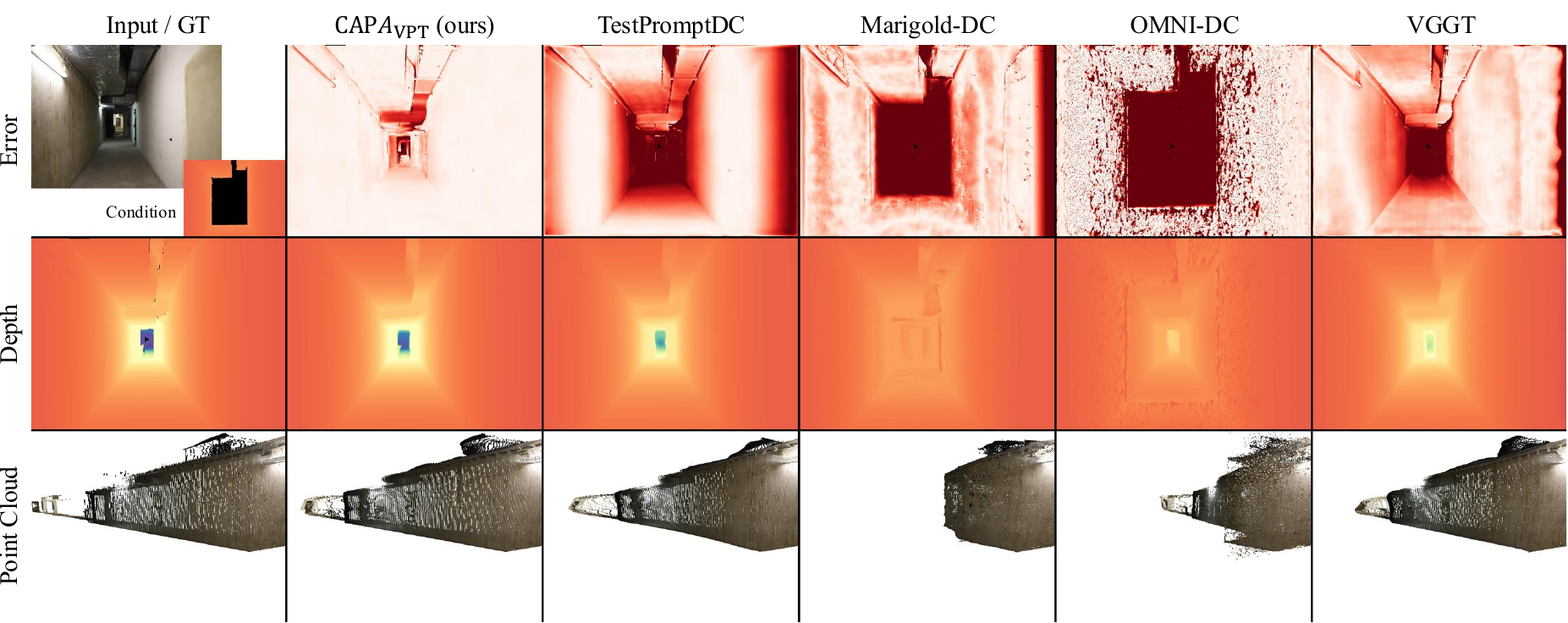}
    \vspace{0.1em} \hrule \vspace{0.1em}
    \includegraphics[width=0.99\linewidth]{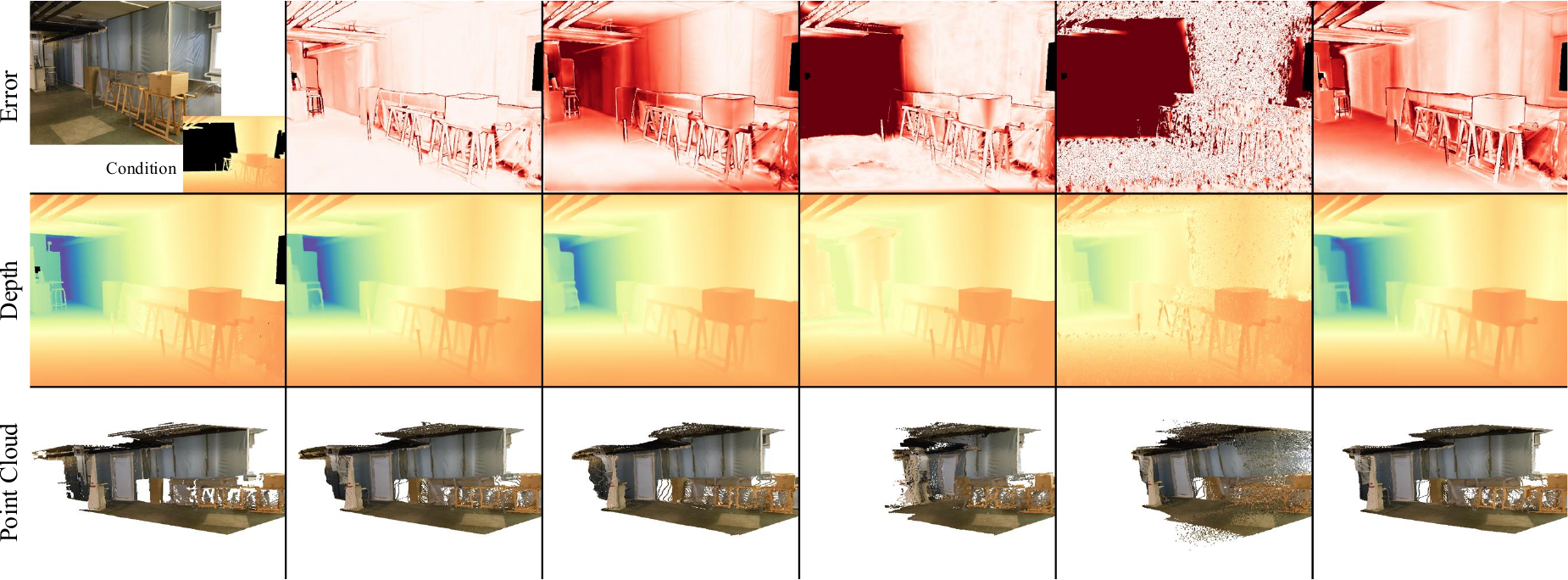}
    \vspace{0.1em} \hrule \vspace{0.1em}
    \includegraphics[width=0.99\linewidth]{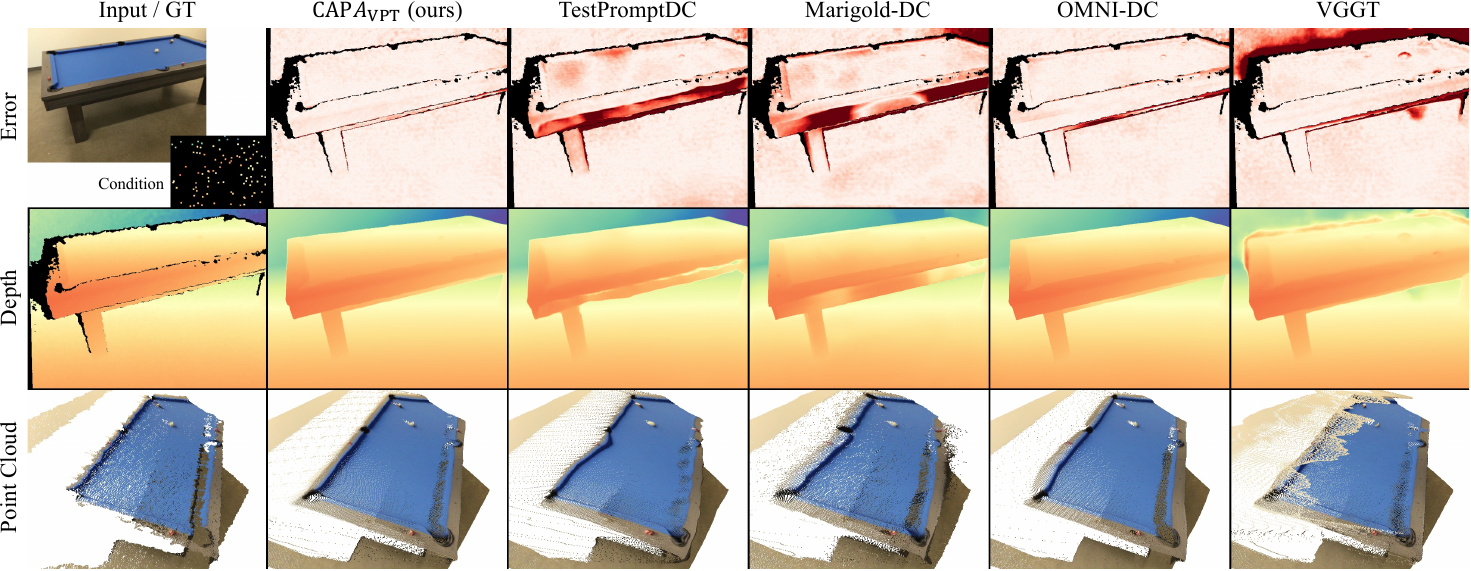}
\end{figure*}

\begin{figure*}[t]
    \vspace{-1em}
    \centering
    \includegraphics[width=0.99\linewidth]{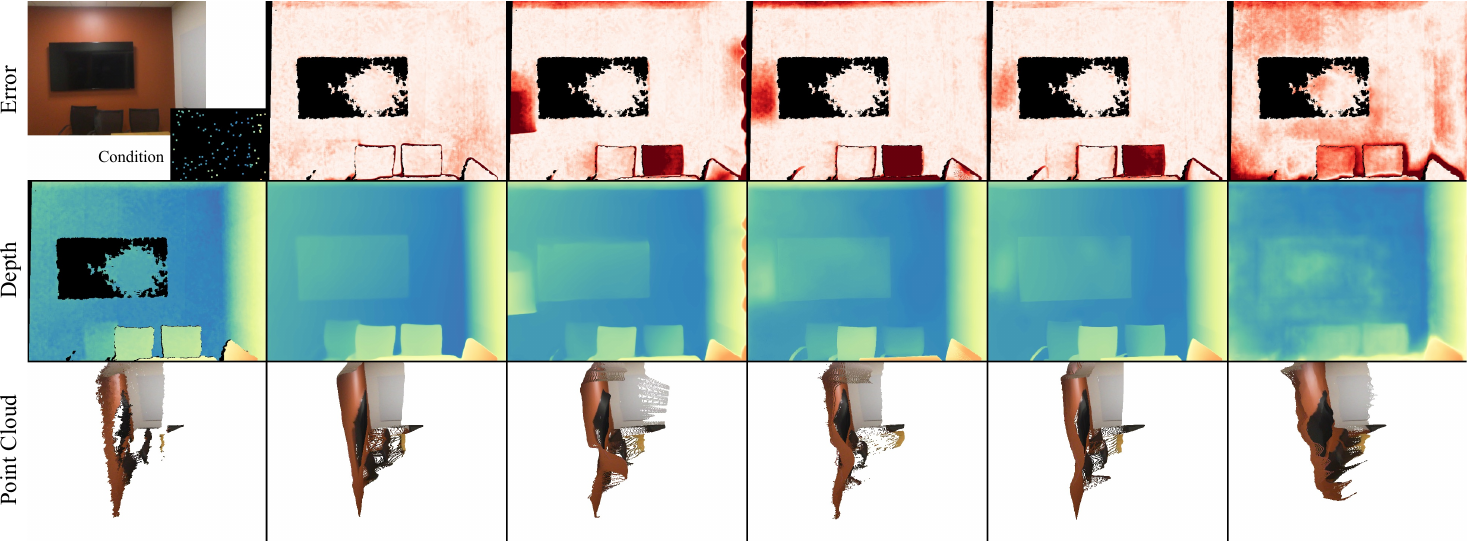}
    \vspace{0.1em} \hrule \vspace{0.1em}
    \includegraphics[width=0.99\linewidth]{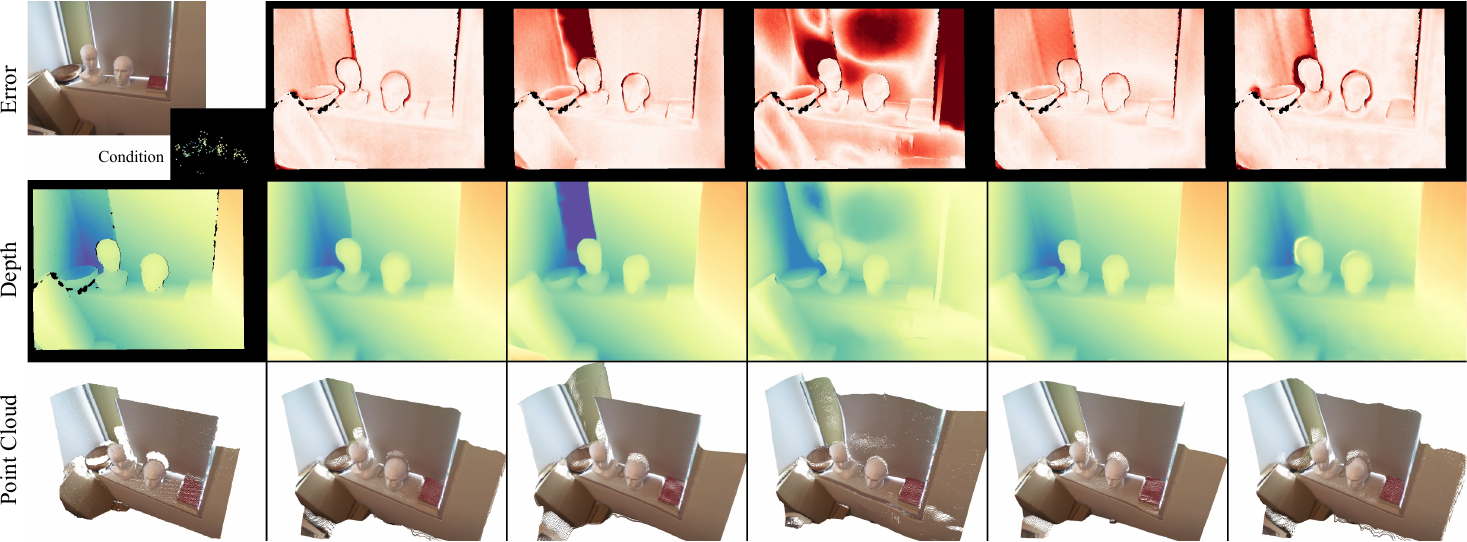}
    \vspace{0.1em} \hrule \vspace{0.1em}
    \includegraphics[width=0.99\linewidth]{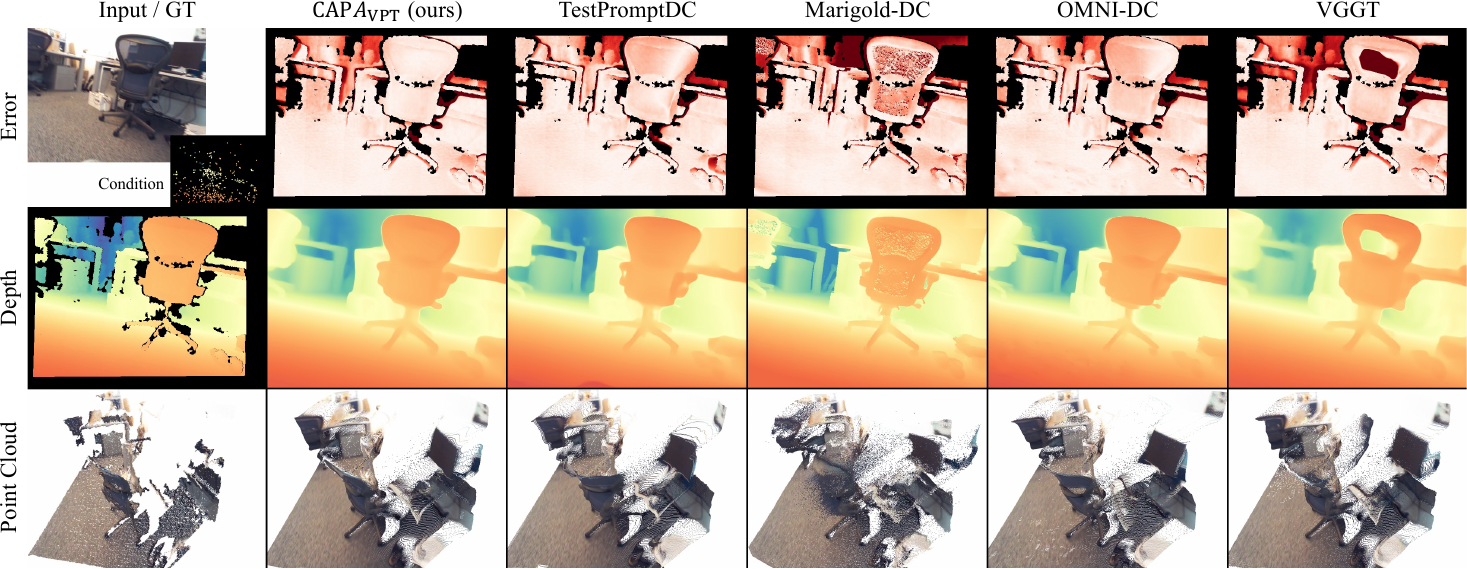}
\end{figure*}
\begin{figure*}[t]
    \vspace{-1em}
    \centering
    \includegraphics[width=0.99\linewidth]{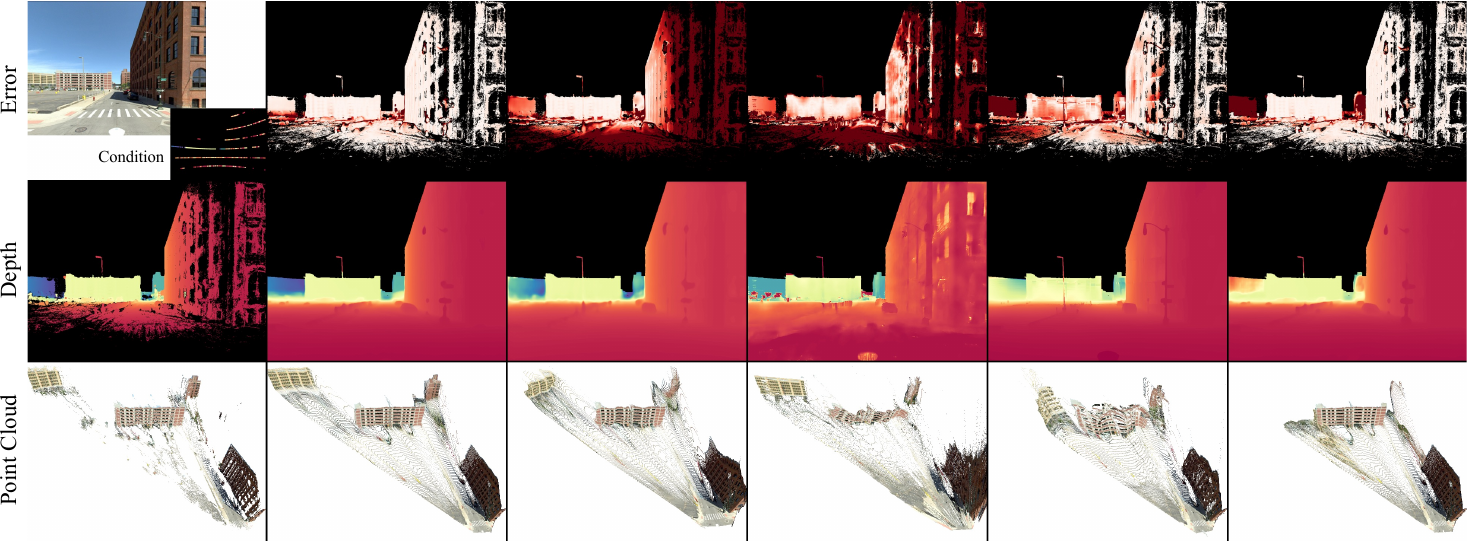}
    \vspace{0.1em} \hrule \vspace{0.1em}
    \includegraphics[width=0.99\linewidth]{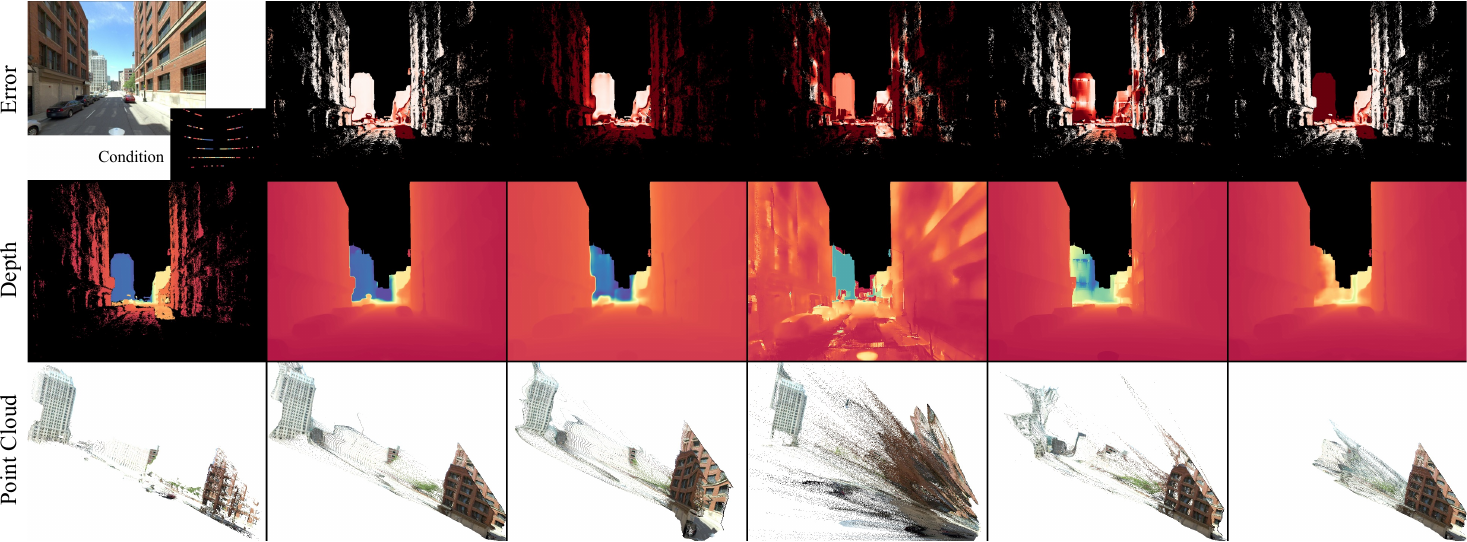}
    \vspace{0.1em} \hrule \vspace{0.1em}
    \includegraphics[width=0.99\linewidth]{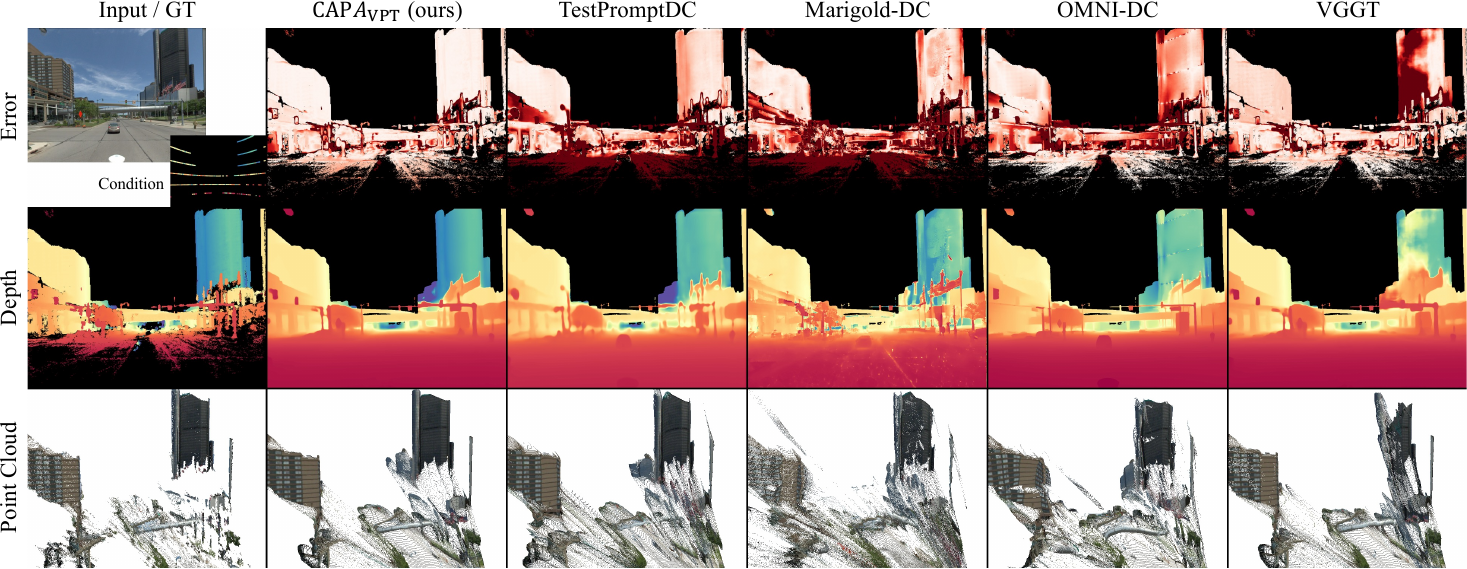}
\end{figure*}

\end{document}